\documentclass[review]{elsarticle}
\pdfoutput=1
\usepackage{lineno,hyperref}
\usepackage{amsfonts} 
\usepackage{fullpage} 
\usepackage{algorithm}
\usepackage{algorithmic}
\usepackage{amsmath}
\usepackage{graphicx} 
\usepackage{mathrsfs}
\usepackage{multirow}
\usepackage{enumitem}

\usepackage{bm}
\usepackage[font=small,skip=1pt]{caption}
\usepackage{subcaption}
\usepackage{amsthm}


\newcommand{\transp}{\mathsf{T}}

\newcommand{\unif}{\operatorname{unif}}

\newcommand{\classlabel}{\operatorname{class\_label}}

\makeatletter
\def\@author#1{\g@addto@macro\elsauthors{\normalsize%
    \def\baselinestretch{1}%
    \upshape\authorsep#1\unskip\textsuperscript{%
      \ifx\@fnmark\@empty\else\unskip\sep\@fnmark\let\sep=,\fi
      \ifx\@corref\@empty\else\unskip\sep\@corref\let\sep=,\fi
      }%
    \def\authorsep{\unskip,\space}%
    \global\let\@fnmark\@empty
    \global\let\@corref\@empty  
    \global\let\sep\@empty}%
    \@eadauthor={#1}
}
\makeatother

\journal{Computational Intelligence for Pattern Recognition}

\begin{document}

\theoremstyle{remark}
\newtheorem{rem}{Remark}

\begin{frontmatter}

\title{Improving Sparse Representation-Based Classification\\
 Using Local Principal Component Analysis}

\author{Chelsea Weaver\corref{cor1}}
\ead{caweaver@math.ucdavis.edu}
\cortext[cor1]{Corresponding author}

\author{Naoki Saito}
\ead{saito@math.ucdavis.edu}

\address{Department of Mathematics\\
University of California, Davis \\
One Shields Avenue\\
Davis, California, 95616, United States}

\begin{abstract}
\emph{Sparse representation-based classification} (SRC), proposed by Wright et al., seeks the sparsest decomposition of a test sample over the dictionary of training samples, with classification to the most-contributing class. Because it assumes test samples can be written as linear combinations of their same-class training samples, the success of SRC depends on the size and representativeness of the training set. Our proposed classification algorithm enlarges the training set by using local principal component analysis to approximate the basis vectors of the tangent hyperplane of the class manifold at each training sample. The dictionary in SRC is replaced by a local dictionary that adapts to the test sample and includes training samples and their corresponding tangent basis vectors. We use a synthetic data set and three face databases to demonstrate that this method can achieve higher classification accuracy than SRC in cases of sparse sampling, nonlinear class manifolds, and stringent dimension reduction. 
\end{abstract}

\begin{keyword}
sparse representation \sep local principal component analysis \sep dictionary learning \sep classification \sep face recognition \sep class manifold 
\end{keyword}

\end{frontmatter}


\section{Introduction}

We are concerned with \emph{classification}, which, in the context of \emph{supervised learning}, is the task of assigning labels to unknown samples given the class information of a training set. It is one of the most important undertakings in pattern recognition and computational intelligence, with applications including the recognition of handwritten digits \cite{lecun:mnist} and face recognition \cite{wri:src,cev:fr,sur:fr}. These tasks are often challenging. For example, in face recognition, the classification algorithm must be robust to within-class variation in properties such as expression, face/head angle, changes in hair or makeup, and differences that may occur in the image environment, most notably, the lighting conditions \cite{sur:fr}. Further, in real-world settings, we must be able to handle greatly-deficient training data (i.e., too few or too similar training samples, in the sense that the given training set is insufficient to generalize the data set's class structure) \cite{ssfr:sur}, as well as occlusion and noise \cite{wri:src}. 

In 2009, Wright et al.\ proposed \emph{sparse representation-based classification} (SRC) \cite{wri:src}. SRC was motivated by the recent boom in the use of sparse representation in signal processing (see, e.g., the work of Cand{\`e}s \cite{can:spa}). The catalyst of these advancements was the discovery that, under certain conditions, the sparsest representation of a signal using an over-complete set of vectors (often called a \emph{dictionary}) could be found by minimizing the $\ell^1$-norm of the representation coefficient vector \cite{don:und}. Since the $\ell^1$-minimization problem is convex, this gave rise to a tractable approach to obtaining the sparsest solution. 

SRC applies this relationship between the minimum $\ell^1$-norm and the sparsest solution to classification. The algorithm seeks the sparsest decomposition of a test sample over the dictionary of training samples via $\ell^1$-minimization, with classification to the class whose corresponding portion of the representation approximates the test sample with least error. The method assumes that class manifolds are linear subspaces, so that the test sample can be represented using training samples in its ground truth class. Wright et al.\ \cite{wri:src} argue that this is precisely the sparsest decomposition of the test sample over the training set. They make the case that sparsity is critical to high-dimensional image classification and that, if properly harnessed, it can lead to superior classification performance, even on highly corrupted or occluded images. Further, good results can be achieved regardless of the choice of image features that are used for classification, provided that the number of retained features is large enough \cite{wri:src}. Though SRC was originally applied to face recognition, similar methods have been employed in clustering \cite{elh:ssc}, dimension reduction \cite{qiao:spp}, and texture and handwritten digit classification \cite{yan:sria}. 

The SRC assumption that class manifolds are linear subspaces is often violated; e.g., facial images that vary in pose and expression are known to lie on nonlinear class manifolds \cite{row:lle,he:lapface}. Additionally, small training set size, one of the primary challenges in face recognition and classification as a whole, can easily make it impossible to represent a given test sample using its same-class training samples, even in the case that the class manifold is linear. However, these reasons alone are not enough to discount SRC even on such data sets, as demonstrated by Wright et al.\ \cite{wri:src} in experiments on the AR face database \cite{AR:face}. AR contains expression and occlusion variations that suggest the underlying class manifolds are nonlinear, yet SRC often outperformed SVM (support vector machines) on AR for a wide variety of feature extraction methods and feature dimensions \cite{wri:src}. To understand how this is possible, consider that SRC decomposes the test sample over the entire training set, and so components of the test sample not within the span of its ground truth class's training samples may be absorbed by training samples from other classes. A similar fail-safe occurs when the class manifolds (linear or otherwise) are sparsely sampled. 

The above discussion, however, illustrates a weakness in SRC. When the algorithm relies on ``wrong-class'' training samples to partially represent or approximate the test sample, misclassification may ensue, especially when the class manifolds are close together. In the case where class manifolds are nonlinear and/or sparsely sampled, so that it is impossible to accurately approximate the test sample using only the training samples in its ground truth class, this approximation could conceivably be improved if we were able to increase the sampling density around the test sample, ``fleshing out'' its local neighborhood on the (correct) class manifold. This is the motivation behind this paper's proposed classification algorithm.

Our contributions in this paper are the following:
\begin{enumerate}[noitemsep,nolistsep]
\item We introduce a classification algorithm that improves SRC by increasing the accuracy and locality of the approximation of the test sample in terms of its ground truth class. Our algorithm is designed to increase the training set via nearby (to the test sample) basis vectors of the hyperplanes approximately tangent to the (unknown) class manifolds. This provides the two-fold benefit of counter-balancing the potential sparse sampling of class manifolds (especially in the case that they are nonlinear) and helping to retain more information in few dimensions when used in conjunction with dimension reduction. 
\item We state guidelines for the setting of parameters in this algorithm and analyze its computational complexity and storage requirements. 
\item We demonstrate that our algorithm leads to classification accuracy exceeding that of traditional SRC and related methods on a synthetic database and three popular face databases. We thoroughly analyze and explain our experimental results (e.g., accuracy, runtime, and dictionary size) of the compared algorithms. 
\item We illustrate that the tangent hyperplane basis vectors used in our method can capture sample details lost during principal component analysis in the case of face recognition. 
\end{enumerate}

Note that both SRC and the method we use to compute the tangent hyperplane basis vectors have previously been proposed. The novelty of the proposed classification algorithm lies in a solid theoretical foundation for combining these two ideas. This motivating foundation is supported empirically---beyond evidence of increased classification accuracy---in experimental results.\footnote{We are referring to Section \ref{sec:pca_tv}.}. Further, by providing thorough guidelines and short-cuts regarding the setting of required parameters, we make it feasible to apply the resulting algorithm in practice.

This paper is organized as follows: In Section \ref{sec:rw}, we discuss work related to our proposed method, and we state SRC in detail in Section \ref{sec:src}. In Section \ref{sec:alg}, we describe our proposed classification algorithm and discuss its parameters, computational complexity, and storage requirements. We present our experimental results in Section \ref{sec:exp_res}, and in Section \ref{sec:fut_work}, we summarize our findings and discuss avenues of future work. 

\textbf{Setup and Notation.} We assume that the input data is represented by vectors in $\mathbb{R}^m$ and that dimension reduction, if used, has already been applied. The training set, i.e., the matrix whose columns are the data samples with known class labels, is denoted by $X_\mathrm{tr} = [\bm{x}_1,\ldots,\bm{x}_{N_\mathrm{tr}}] \in \mathbb{R}^{m\times N_{\mathrm{tr}}}$. The number of classes is denoted by $L \in \mathbb{N}$, and we assume that there are $N_l$ training samples in class $l$, $1\leq l \leq L$. Lastly, we refer to a given test sample by $\bm{y} \in \mathbb{R}^m$.

\section{Related Work} \label{sec:rw}

The approach of using tangent hyperplanes for pattern recognition is not new. When the data is assumed to lie on a low-dimensional manifold, local tangent hyperplanes are a simple and intuitive approach to enhancing the data set and gaining insight into the manifold structure. Our proposed method is very much related to \emph{tangent distance classification} (TDC) \cite{sim:tdc,cha:tdc,yan:ltd}, which constructs local tangent hyperplanes of the class manifolds, computes the distances between these hyperplanes and the given test sample, and then classifies the test sample to the class with the closest hyperplane. We show in Section \ref{sec:exp_res} that our proposed method's integration of tangent hyperplane basis vectors into the sparse representation framework generally outperforms TDC.

On the other hand, approaches to address the limiting linear subspace assumption (i.e., the assumption that class manifolds are linear subspaces) in SRC have been proposed. For example, Ho et al.\ extended sparse coding and dictionary learning to general Riemannian manifolds \cite{xie:nlsrc}. Admittedly only a first step in meeting their ultimate objective, Ho et al.'s work requires explicit knowledge of the class manifolds. This is an unsatisfiable condition in many real-world classification problems and is not a requirement of our proposed algorithm. Alternatively, \emph{kernel methods} have been effective in overcoming SRC's linearity assumption, as nonlinear relationships in the original space may be linear in kernel space given an appropriate choice of kernel \cite{yin:ksrc}.
 
Several ``local'' modifications of SRC implicitly ameliorate the linearity assumption; in \emph{collaborative neighbor representation-based classification} \cite{waq:cnrc} and \emph{locality-sensitive dictionary learning} (LSDL-SRC) \cite{wei:lsdl}, for instance, coefficients of the representation are constrained by their corresponding training samples' distances to the test sample, and so these algorithms need only assume linearity at the local level. Our proposed method is designed to improve not only the locality but also the accuracy of the approximation of the test sample in terms of its ground truth class. Section \ref{sec:exp_res} contains an experimental comparison between our proposed method and LSDL-SRC, as well as a discussion thereof.

Other classification algorithms have been proposed that are similar to ours in that they aim to enlarge or otherwise enhance the training set in SRC. Such methods for face recognition, for example, include the use of virtual images that exploit the symmetry of the human face, as in both the method of Xu et al.\ \cite{xu:mir} and \emph{sample pair based sparse representation classification} \cite{zha:spsrc}. Though visual comparison of these virtual images and our recovered tangent vectors (see Section \ref{sec:pca_tv}) could be informative, our proposed method can be used for general classification.

Additionally, there have been many local modifications to the sparse representation framework with objectives other than classification. For example, Li et al.'s \emph{robust structured subspace learning} (RSSL) \cite{li:rssl} uses the $\ell_{2,1}$-norm for sparse feature extraction, combining high-level semantics with low-level, locality-preserving features. In the feature selection algorithm \emph{clustering-guided sparse structural learning} (CGSSL) by Li et al.\ \cite{li:clust}, features are jointly selected using sparse regularization (via the $\ell_{2,1}$-norm) and a non-negative spectral clustering objective. Not only are the selected features sparse; they also are the most discriminative features in terms of predicting the cluster indicators in both the original space and a lower-dimensional subspace on which the data is assumed to lie.

\section{Sparse Representation-Based Classification} \label{sec:src}

SRC \cite{wri:src} solves the optimization problem
\begin{equation}\label{eq:src_opt}
\bm{\alpha}^* := \arg \min_{\bm{\alpha}\in \mathbb{R}^{N_\mathrm{tr}}} \|\bm{\alpha}\|_1, \text{ subject to } \bm{y} = X_\mathrm{tr} \bm{\alpha}.
\end{equation}

It is assumed that the training samples have been normalized to have $\ell^2$-norm equal to $1$, so that the representation in Eq.~\eqref{eq:src_opt} will not be affected by the samples' magnitudes. The use of the $\ell^1$-norm in the objective function is designed to approximate the $\ell^0$-``norm,'' i.e., to aim at finding the smallest number of training samples that can accurately represent the test sample $\bm{y}$. It is argued that the nonzero coefficients in the representation will occur primarily at training samples in the same class, so that
\begin{equation} \label{eq:src_class}
\classlabel(\bm{y}) = \arg \min_{1\leq l \leq L} \big\|\bm{y} - X_\mathrm{tr}\delta_l(\bm{\alpha}^*)\big\|_2
\end{equation}
produces the correct class assignment. Here, $\delta_l$ is the indicator function that acts as the identity on all coordinates corresponding to samples in class $l$ and sets the remaining coordinates to zero. In other words, $\bm{y}$ is assigned to the class whose training samples contribute the most to the sparsest representation of $\bm{y}$ over the entire training set.

The reasoning behind this is the following: It is assumed that the class manifolds are linear subspaces, so that if each class's training set contains a spanning set of the corresponding subspace, the test sample can be expressed as a linear combination of training samples in its ground truth class. If the number of training samples in each class is small relative to the number of total training samples $N_\mathrm{tr}$, this representation is naturally sparse \cite{wri:src}. 

As real-world data is often corrupted by noise, the constrained $\ell^1$-minimization problem in Eq.~\eqref{eq:src_opt} may be replaced with its regularized version
\begin{equation} \label{eq:src_opt_noise}
\bm{\alpha}^* := \arg \min_{\bm{\alpha}\in \mathbb{R}^{N_\mathrm{tr}}} \Big \{\frac{1}{2} \|\bm{y}-X_\mathrm{tr}\bm{\alpha}\|_2^2 + \lambda\|\bm{\alpha}\|_1\Big \}.
\end{equation}
Here, $\lambda$ is the trade-off between error in the approximation and the sparsity of the coefficient vector. We summarize SRC in Algorithm 1.

\begin{algorithm}
\caption{Sparse Representation-Based Classification (SRC) \cite{wri:src}}
\label{alg:src}
\begin{algorithmic}[1]
\REQUIRE Matrix of training samples $X_\mathrm{tr} \in \mathbb{R}^{m\times N_{\mathrm{tr}}}$; test sample $\bm{y} \in \mathbb{R}^m$; number of classes $L$; and error/sparsity trade-off $\lambda$ (optional)
\ENSURE The computed class label of $\bm{y}$: $\classlabel(\bm{y})$
\STATE Normalize each column of $X_\mathrm{tr}$ to have $\ell^2$-norm equal to $1$.
\STATE Use an $\ell^1$-minimization algorithm to solve either the constrained problem \eqref{eq:src_opt} or the regularized problem \eqref{eq:src_opt_noise}.
\FOR{each class $l=1,\ldots, L$,}
\STATE Compute the norm of the class $l$ residual:
$\mathrm{err}_l(\bm{y}) := \big\|\bm{y} - X_\mathrm{tr}\delta_l(\bm{\alpha}^*)\big\|_2$.
\ENDFOR
\STATE Classify the test sample $\bm{y}$ according to $\classlabel({\bm{y}}) = \arg\min_{1\leq l \leq L} \{\mathrm{err}_l(\bm{y})\}$.
\end{algorithmic}
\end{algorithm}

\begin{rem}
We briefly note that, in the case that some classes contain very few samples, SRC is not a good candidate for \emph{oversampling}, or using repeated training samples to even out the class count. This is because the linear span of the training samples is invariant to the addition of repeat samples and the classification result will be unaffected. Thus there is no obvious solution to dealing with undersampled classes in SRC.
\end{rem}

\section{Proposed Algorithm}\label{sec:alg}

\subsection{Local Principal Component Analysis Sparse Representation-Based Classification}

Our proposed algorithm, \emph{local principal component analysis sparse representation-based classification} (LPCA-SRC), is essentially SRC with a modified dictionary. This dictionary is constructed in two steps: (i) an offline phase, and (ii) an online phase.
 
In the offline phase of the algorithm, we generate new training samples as a means of increasing the sampling density. Instead of the linear subspace assumption in SRC, we assume that class manifolds are well-approximated by local tangent hyperplanes. To generate new training samples, we approximate these tangent hyperplanes at individual training samples using \emph{local principal component analysis} (local PCA), and then add the basis vectors of these tangent hyperplanes (after randomly-scaling and shifting them as described in Step \ref{scale} of Algorithm \ref{lpca_src_offline} and explained in Section \ref{sec:pruning}) to the original training set. Naturally, the shifted and scaled tangent hyperplane basis vectors (hereon referred to as ``tangent vectors'') inherit the labels of their corresponding training samples. The result is an amended dictionary over which a generic test sample can ideally be decomposed using samples that approximate a local patch on the correct class manifold. In the case that the class manifolds are sparsely sampled and/or nonlinear, this allows for a more accurate approximation of $\bm{y}$ using training samples (and their computed tangent vectors) from the test sample's ground truth class. Even in the case that class manifolds are linear subspaces, this technique ideally increases the sampling density around $\bm{y}$ on its (unknown) class manifold so that it may be expressed in terms of \emph{nearby} samples.

In the online phase of LPCA-SRC, this extended training set is ``pruned'' relative to the given test sample, increasing computational efficiency and the locality of the resulting dictionary. Training samples (along with their tangent vectors) are eliminated from the dictionary if their Euclidean distances to the given test sample are greater than a threshold, and then classification proceeds as in SRC as the test sample is sparsely decomposed (via $\ell^1$-minimization) over this local dictionary.

The method in LPCA-SRC has an additional benefit: When SRC is applied to the classification of high-resolution images (e.g., $>O(10^4)$ pixels), some method of dimension reduction is generally necessary to reduce the dimension of the raw samples, due to the high computational complexity of solving the $\ell^1$-minimization problem. Basic dimension reduction methods, such as \emph{principal component analysis} (PCA), may result in the loss of class-discriminating details when the PCA feature dimension is small. In Section \ref{sec:pca_tv}, we show that the tangent vectors computed in LPCA-SRC can contain details of the raw images that have been lost in the dimension reduction process. 

\begin{rem}
This remark serves to draw a distinction between our use of sparse representation and local PCA for classification, and our use of (non-local) PCA to pre-process data samples prior to classification in some of our experiments. Sparse representation and local PCA can themselves be used (separately) for dimension reduction; see, for example, the papers of Qiao et al.\ \cite{qiao:spp} and Kambhatla and Leen \cite{kam:lpca}. We stress that dimension reduction is not the subject of this paper, and in fact we use neither sparse representation nor local PCA to accomplish this task at any point. Instead, we focus on classification and in integrating sparse representation and local PCA towards this purpose. When dimension reduction is used in Section \ref{sec:face}, we use (non-local) PCA simply as a means of pre-processing the data before image classification. 
\end{rem}

We formally state the offline and online portions of our proposed algorithm in Algorithms \ref{lpca_src_offline} and \ref{lpca_src_online}, respectively. Obviously, by the definition of ``offline phase,'' the tangent vectors need only be computed once for any number of test samples. More details regarding the user-set parameters $d$, $n$ and $\lambda$ are provided in Sections \ref{sec:pars_dn} and \ref{sec:par_set}, and an explanation of the pruning parameter $r$ and the tangent vector scaling factor $c$ (in Step \ref{scale_step} of Algorithm \ref{lpca_src_offline}) are given in Section \ref{sec:pruning}.

\begin{figure}[!htb]
\centering
\includegraphics[width=\linewidth]{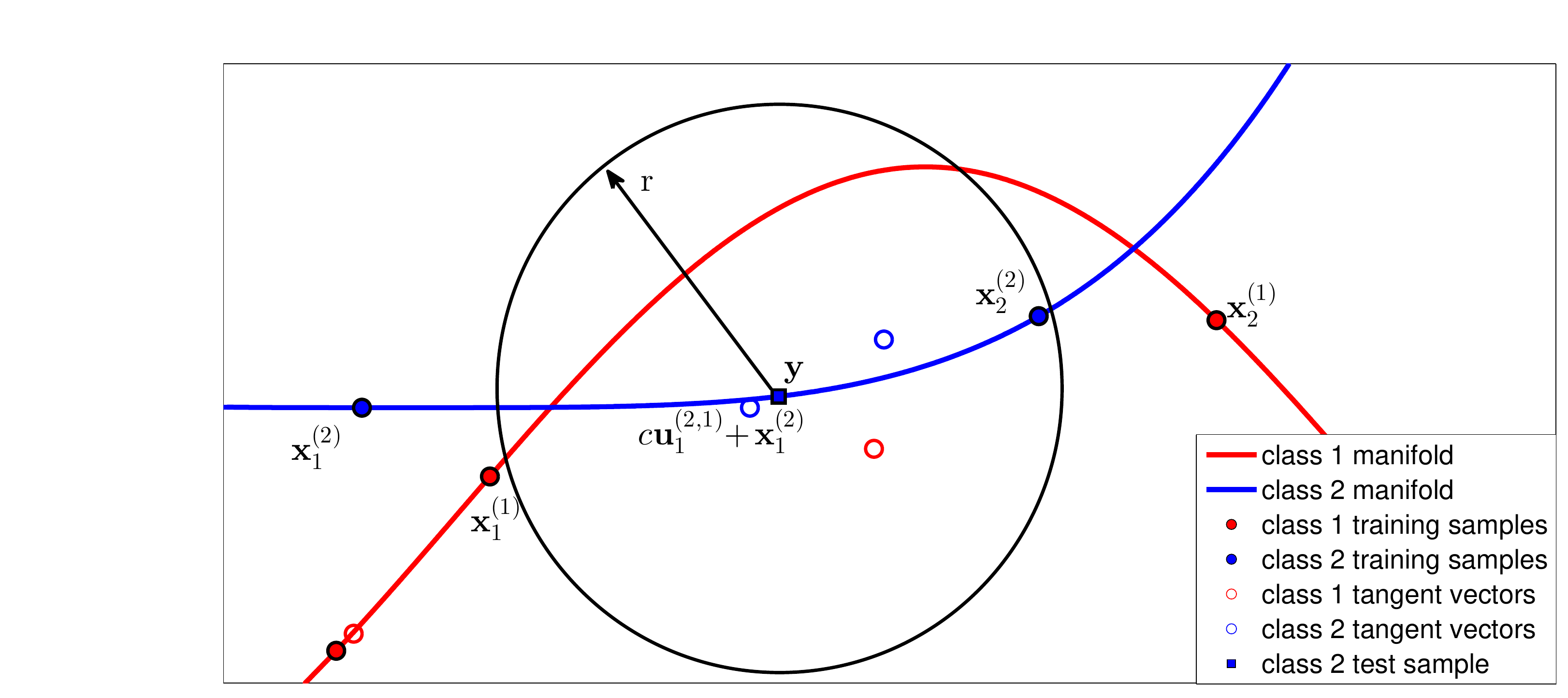}
\caption{An example use of LPCA-SRC's tangent vectors and pruning parameter in the SRC framework. Only training samples and tangent vectors relevant to classification of the test sample $\bm{y}$ have been labeled.}
\label{fig:LPCA_SRC_illustration}
\end{figure}

Figure \ref{fig:LPCA_SRC_illustration} illustrates the efficacy of LPCA-SRC's tangent vectors and pruning parameter in the sparse representation framework. The figure shows two classes, represented by the colors red and blue. The training samples in each class are represented by solid colored circles. There is one test sample $\bm{y}$ displayed in the figure, a member of class 2 (blue) and depicted by a solid blue square. Observe that, before the use of tangent vectors, $\bm{y}$ is closer to the subspace\footnote{Technically speaking, we are referring to the affine subspace in this illustration; In SRC, instead the \emph{linear} subspace is used. We have tweaked the algorithm slightly to be able to demonstrate an example in low dimension.} spanned by $\bm{x}_1^{(1)}$ and $\bm{x}_2^{(1)}$ (which are class 1 training samples) than the subspace spanned by the class 2 training samples $\bm{x}_1^{(2)}$ and $\bm{x}_2^{(2)}$. Thus $\bm{y}$ would be incorrectly classified by SRC in this scenario.

After the addition of tangent vectors (which are represented by unfilled circles), in particular, the class 2 tangent vector $c\bm{u}_1^{(2,1)} + \bm{x}_1^{(2)}$, $\bm{y}$ is closest to the subspace generated by this tangent vector and $\bm{x}_2^{(2)}$. Thus the test sample would be correctly classified by LPCA-SRC in this scenario.

The use of the pruning parameter $r$ independently avoids the problem of misclassification. If we consider only samples in the local neighborhood of $\bm{y}$ (contained in the circle of radius $r$), the misleading class 1 samples $\bm{x}_1^{(1)}$ and $\bm{x}_2^{(1)}$ are eliminated from consideration, leading to the correct classification of $\bm{y}$.

Thus these two mechanisms in LPCA-SRC---its use of tangent vectors and its localizing pruning parameter---make it especially designed to succeed in these cases of sparse sampling and nonlinear class manifolds in which SRC fails.

\begin{algorithm}
\caption{Local PCA Sparse Representation-Based Classification (LPCA-SRC): \textbf{OFFLINE PHASE}}
\label{lpca_src_offline} 
\begin{algorithmic}[1]
\REQUIRE $X_\mathrm{tr} = [\bm{x}_1 \ldots, \bm{x}_{N_{\mathrm{tr}}}] \in \mathbb{R}^{m\times N_\mathrm{tr}}$; number of classes $L$; local PCA parameters $d$ (estimate of class manifold dimension) and $n$ (number of neighbors)
\ENSURE The normalized extended dictionary $D \in \mathbb{R}^{m\times (N_\mathrm{\textnormal{tr}}(d+1))}$; pruning parameter $r$
\STATE Normalize the columns of $X_\mathrm{tr}$ to have $\ell^2$-norm equal to $1$.
\FOR{each class $l=1,\ldots,L$}
\STATE Let $\mathcal{X}^{(l)}$ be the set of class $l$ training samples contained in $X_\mathrm{tr}$. 
\FOR{each class $l$ training sample $\bm{x}_i^{(l)}$, $i=1, \ldots,N_l$}
\STATE Approximate the tangent hyperplane of the $l$th class manifold at $\bm{x}_i^{(l)}$ as follows:\label{lpca_alg2} 
\begin{itemize}[noitemsep,nolistsep]
\item Use local PCA in Algorithm \ref{alg1} with set of samples $\mathcal{X}^{(l)}$ (the samples in the $l$th class), selected sample $\bm{x}_i^{(l)}$, and parameters $d$ and $n$ to compute a basis $U^{(l,i)} := [\bm{u}_1^{(l,i)},\ldots,\bm{u}_{d}^{(l,i)}]$ of an approximate tangent hyperplane at $\bm{x}_i^{(l)}$ along class $l$.
\item Store the basis $U^{(l,i)}$ and the quantity $r_i^{(l)} := \| \bm{x}_{i_{n+1}}^{(l)}-\bm{x}_i^{(l)} \|_2$, the distance between $\bm{x}_i^{(l)}$ and its $(n+1)$st nearest neighbor in the $l$th class. 
\end{itemize}
\ENDFOR
\ENDFOR
\STATE Define the pruning parameter $r := \text{median} \big\{ r_i^{(l)}\, | \, 1\leq i \leq N_l, \, 1\leq l \leq L \big\}$.
\STATE Initialize the extended dictionary $D = \emptyset$. 
\FOR{each class $l=1,\ldots,L$}
\FOR{each class $l$ training sample $\bm{x}_i^{(l)}$, $i=1, \ldots,N_l$}
\STATE\label{scale_step} Set $c:= r \gamma$, $\gamma \sim \unif(0,1)$, and form $\tilde{X}^{(l,i)} := \big[c\bm{u}_1^{(l,i)} + \bm{x}_i^{(l)},\ldots,c\bm{u}_{d}^{(l,i)} + \bm{x}_i^{(l)}, \bm{x}_i^{(l)}\big] \in \mathbb{R}^{m\times(d+1)}$. \label{scale}
\STATE Normalize the columns of $\tilde{X}^{(l,i)}$ to have $\ell^2$-norm equal to $1$ and add it to the extended dictionary: \\
$D = [D,\tilde{X}^{(l,i)}]$.
\ENDFOR
\ENDFOR
\end{algorithmic}
\end{algorithm}

\begin{algorithm}
\caption{Local PCA Sparse Representation-Based Classification (LPCA-SRC): \textbf{ONLINE PHASE}}
\label{lpca_src_online} 
\begin{algorithmic}[1]
\REQUIRE Test sample $\bm{y} \in \mathbb{R}^m$; normalized extended dictionary $D$; pruning parameter $r$; estimate of class manifold dimension $d$; error/sparsity trade-off $\lambda$ (optional)
\ENSURE The computed class label of $\bm{y}$: $\classlabel(\bm{y})$. 
\STATE Normalize $\bm{y}$ to have $\|\bm{y}\|_2 = 1$.
\STATE Initialize the pruned dictionary $D_{\bm{y}} = \emptyset$ and set $N_{\bm{y}} = 0$ (\# of columns of $D_{\bm{y}}$).
\FOR{each class $l = 1,\ldots, L$}
\FOR{each class $l$ training sample $\bm{x}_i^{(l)}$, $i=1,\ldots,N_l$}
\IF{$\big\|\bm{y} - \bm{x}_i^{(l)}\big\|_2 \leq r$ or $\big\|\bm{y} - (-\bm{x}_i^{(l)})\big\|_2 \leq r$} 
\STATE Add the portion $\tilde{X}^{(l,i)}$ of $D$ corresponding to $\bm{x}_i^{(l)}$ and its tangent vectors to the pruned dictionary: $D_{\bm{y}} = [D_{\bm{y}},\tilde{X}^{(l,i)}]$. Assign the columns of $\tilde{X}^{(l,i)}$ class $l$ labels. Update $N_{\bm{y}} = N_{\bm{y}} + (d+1)$.
\ENDIF
\ENDFOR
\ENDFOR
\STATE Use an $\ell^1$-minimization algorithm to compute the solution to the constrained problem
\begin{equation} \label{eq:l1_exact}
\bm{\alpha}^* := \arg \min_{\bm{\alpha}\in \mathbb{R}^{N_{\bm{y}}}} \big \{ \|\bm{\alpha}\|_1 \text{ s.t. } \bm{y} = D_{\bm{y}} \bm{\alpha}\big\}
\end{equation}
or the regularized problem
\begin{equation} \label{eq:l1_approx}
\bm{\alpha}^* := \arg \min_{\bm{\alpha}\in \mathbb{R}^{N_{\bm{y}}}} \Big \{ \frac{1}{2}\|\bm{y}-D_{\bm{y}} \bm{\alpha}\|_2^2+ \lambda\|\bm{\alpha}\|_1\Big\}.
\end{equation}
\FOR{each class $l=1,\ldots, L$,}
\STATE Compute the norm of the class $l$ residual: 
$\mathrm{err}_l(\bm{y}) := \big\|\bm{y} - D_{\bm{y}}\delta_l(\bm{\alpha}^*)\big\|_2$.
\ENDFOR
\STATE Classify the test sample $\bm{y}$ according to $\classlabel({\bm{y}}) = \arg\min_{1\leq l \leq L} \{\mathrm{err}_l(\bm{y})\}$.
\end{algorithmic}
\end{algorithm}

\subsection{Local Principal Component Analysis}

In LPCA-SRC (in particular, Step \ref{lpca_alg2} of Algorithm \ref{lpca_src_offline}), we use the local PCA technique of Singer and Wu \cite{sin:vdm} to compute the tangent hyperplane basis $U^{(l,i)}$. We outline our implementation of their method in Algorithm \ref{alg1}. It computes a basis for the tangent hyperplane $T_{\bm{x}_i}\mathscr{M}$ at a point $\bm{x}_i$ on the manifold $\mathscr{M}$, where it is assumed that the local neighborhood of $\bm{x}_i$ on $\mathscr{M}$ can be well-approximated by a tangent hyperplane of some dimension $d<m$. A particular strength of Singer and Wu's method is the weighting of neighbors by their Euclidean distances to the point $\bm{x}_i$, so that closer neighbors play a more important role in the construction of the local tangent hyperplane. 

\begin{figure}[!htb]
\begin{center}
\includegraphics[width=\linewidth]{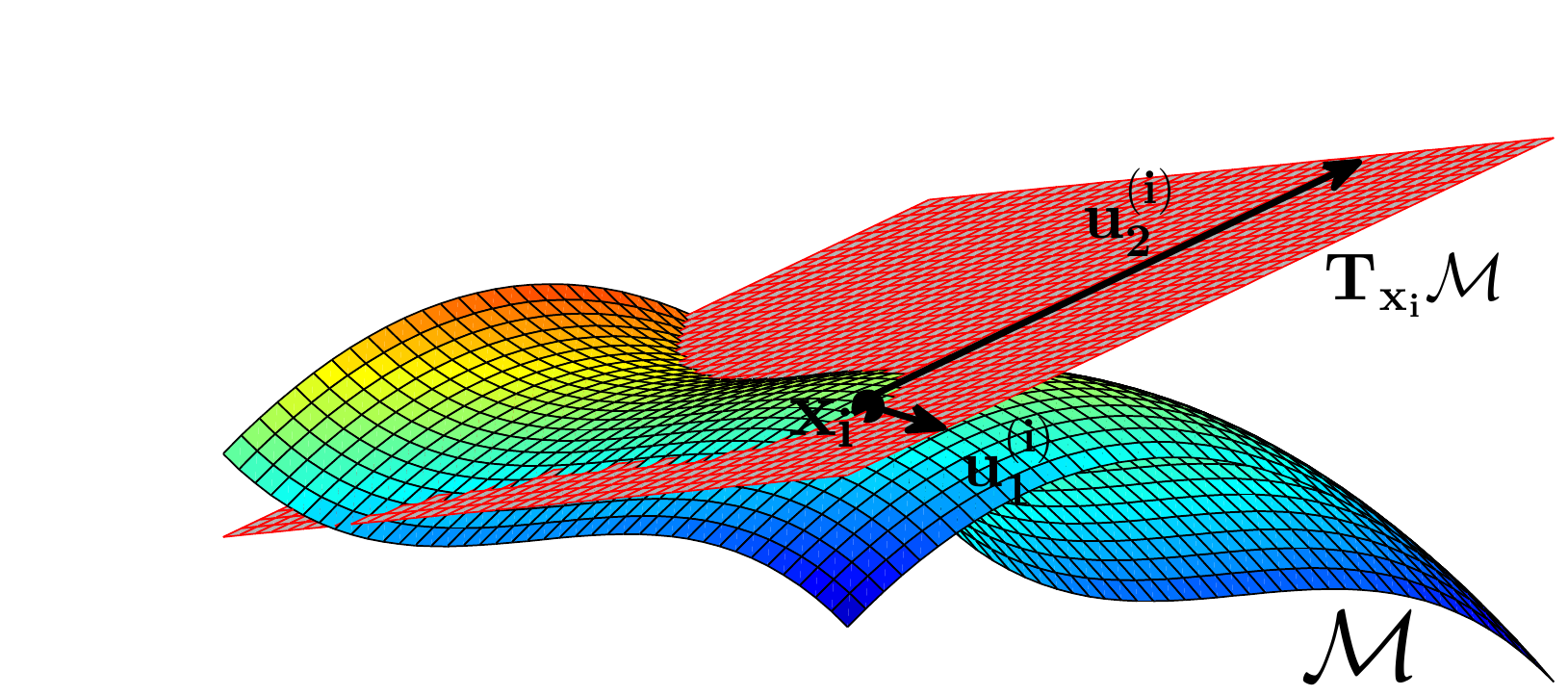}
\caption{An illustration of the tangent plane and the tangent basis vectors $\bm{u}_1^{(i)}$ and $\bm{u}_2^{(i)}$at the sample $\bm{x}_i$ on the manifold $\mathscr{M}$. Here, the intrinsic dimension is $d=2$. }
\label{fig:tv_illustration}
\end{center}
\end{figure}

A simple illustration of the tangent basis vectors found using local PCA is shown in Figure \ref{fig:tv_illustration}.

\begin{algorithm}
\caption{Local Principal Component Analysis (Local PCA, adapted from Singer and Wu \cite{sin:vdm})}
\label{alg1}
\begin{algorithmic}[1]
\REQUIRE Set of samples $\mathcal{X}$; selected sample $\bm{x}_i \in \mathcal{X}$; dimension of tangent hyperplane $d$; number of neighbors $n$
\ENSURE The basis $U^{(l,i)}$ of the approximated tangent hyperplane at the point $\bm{x}_i$
\STATE Find the $n+1$ nearest neighbors (with respect to Euclidean distance) of $\bm{x}_i$ in $\mathcal{X}\backslash \bm{x}_i$. Store the $n$ nearest neighbors as columns of the matrix $X_i := [\bm{x}_{i_1},\ldots,\bm{x}_{i_n}]$ and use the $(n+1)$st nearest neighbor to define 
$\epsilon_\mathrm{PCA} := \|\bm{x}_{i_{n+1}} - \bm{x}_i\|_2^2$.
\STATE Form the matrix $\overline{X}_i$ by centering the columns of $X_i$ around $\bm{x}_i$: $\overline{X}_i := [\bm{x}_{i_1}-\bm{x}_i,\ldots, \bm{x}_{i_n}-\bm{x}_i]$.
\STATE Form a diagonal weight matrix $D_i$ based on the distance between each neighbor and $\bm{x}_i$ as follows: Let \\
$D_i(j,j) = \sqrt{K\Big(\frac{\|\bm{x}_{i_j}-\bm{x}_i\|_2}{\sqrt{\epsilon_\mathrm{PCA}}}\Big)}$, $j=1,\ldots,n$, where $K$ is the Epanechnikov kernel given by \\
$K(u) := (1-u^2)\bm{\chi}_{[0,1]}$.
\STATE Form the weighted matrix $B_i :=\overline{X}_iD_i$.
\STATE Find the first $d$ left singular vectors of $B_i$ using singular value decomposition. Denote these vectors by $\bm{u}_1^{(i)},\ldots,\bm{u}_d^{(i)}$.
\end{algorithmic}
\end{algorithm}

\subsection{Remarks on Parameters} \label{sec:pars}

In this subsection, we detail the roles of the parameters in LPCA-SRC and suggest strategies for estimating those that must be determined by the user. 

\subsubsection{Estimate of class manifold dimension and number of neighbors} \label{sec:pars_dn}

Recall that $d$ is the estimated dimension of each class manifold and $n$ is the number of neighbors used in local PCA. Both parameters must be inputted by the user in our proposed algorithm. The number of samples in the smallest training class, denoted $N_{l_\mathrm{min}}$, limits the range of values for $d$ and $n$ that may be used. Specifically, 
\begin{equation} \label{eq:dn_range}
1 \leq d \leq n < N_{l_\mathrm{min}}-1. 
\end{equation}
This follows from the fact that each training sample must have at least $n+1$ neighbors in its own class, with the dimension $d$ of the tangent hyperplane being bounded above by the number of columns $n$ in the weighted matrix of neighbors $B_i$. It is important to observe that when the classes are small (as is often the case in face recognition), there are few options for the values of $d$ and $n$ per Eq.~\eqref{eq:dn_range}. Thus these parameters may be efficiently set using cross-validation. This was the method we used to set $d$ and $n$ in the experiments in Section \ref{sec:exp_res}. We discuss a recommended cross-validation procedure in Section \ref{sec:par_set}.

\begin{rem} \label{rem:d}
Interestingly, when cross-validation is used to set $d$, we find empirically that $d$ is often selected to be smaller than the (expected) true class manifold dimension. Further, in these cases, increasing $d$ from the selected value (i.e., increasing the number of tangent vectors used) does not significantly increase classification accuracy. We expect that the addition of even a small number of tangent vectors (those indicating the directions of maximum variance on their local manifolds, per the local PCA algorithm) is enough to improve the approximation of the test sample in terms of its ground truth class. Additional tangent vectors are often unneeded. Since the value of $d$ largely affects LPCA-SRC's computational complexity and storage requirements, these observations suggest that when the true manifold dimension is large, it is better to underestimate it than overestimate it. Further, setting $d=1$ can often produce a good result, hence $d=1$ could be used by default.

There are other methods for determining $d$ besides cross-validation and fixing $d=1$. One may use the multiscale SVD algorithm of Little et al.\ \cite{mag:msvd} or Ceruti et al.'s DANCo (\emph{Dimensionality from Angle and Norm Concentration} \cite{cer:dan}). However, in our experiments in Section \ref{sec:exp_res}, we set $d$ using cross-validation. See Section \ref{sec:par_set} below.
\end{rem}

\begin{rem}
Certainly, the parameters $d$ and $n$ could vary per class, i.e., $d$ and $n$ could be replaced with $d_l$ and $n_l$, respectively, for $l=1,\ldots,L$. In face recognition, however, if each subject is photographed under similar conditions, e.g., the same set of lighting configurations, then we expect that the class manifold dimension is approximately the same for each subject. Further, without some prior knowledge of the class manifold structure, using distinct $d$ and $n$ for each class may unnecessarily complicate the setting of parameters in LPCA-SRC.
\end{rem}

\subsubsection{Using cross-validation to set multiple parameters} \label{sec:par_set}

On data sets of which we have little prior knowledge, it may be necessary to use cross-validation to set multiple parameters in LPCA-SRC. Since grid search (searching through all parameter combinations in a brute-force manner) is typically expensive, we suggest that cross-validation be applied to the parameters $n$, $\lambda$, and $d$, consecutively in that order as needed.\footnote{If the constrained optimization problem (Eq.~\eqref{eq:l1_exact}) is used, the error/sparsity trade-off $\lambda$ is not needed.} During this process, we recommend holding the error/sparsity trade-off $\lambda$ (if used) equal to a small, positive value (e.g., $\lambda=0.001$) and setting $d=1$ until these parameters' respective values are determined. We justify and detail this approach below.

Our reasons for suggesting this consecutive cross-validation procedure is the following: During experiments, we found that the LPCA-SRC algorithm can be quite sensitive to the setting of $n$, especially when there are many samples in each training class (since there are many possible values for $n$). This is expected, as the setting of $n$ affects both the accuracy of the tangent vectors and the pruning parameter $r$. In contrast, LPCA-SRC is empirically fairly robust to the values of $\lambda$ and $d$ used, and as mentioned in Remark \ref{rem:d}, setting $d=1$ can result in quite good performance in LPCA-SRC, even when the true dimension of the class manifolds is expected to be larger.

\subsubsection{Pruning parameter} \label{sec:pruning}

First, we stress that the pruning parameter $r$ is not a user-set parameter. Its value is automatically computed in the offline phase of LPCA-SRC (Algorithm \ref{lpca_src_offline}). We explain this process here.

Recall that we only include a training sample $\bm{x}_i^{(l)}$ and its tangent vectors in the pruned dictionary $D_{\bm{y}}$ if $\bm{x}_i^{(l)}$ (or its negative) is in the closed Euclidean ball $\overline{B_m(\bm{y}, r)} \subset \mathbb{R}^m$ with center $\bm{y}$ and radius $r$. Thus $r$ is a parameter that prunes the extended dictionary $D$ to obtain $D_{\bm{y}}$. A smaller dictionary is good in terms of computational complexity, as the $\ell^1$-minimization algorithm will run faster. Further, we can obtain this computational speedup without (theoretically) degrading classification accuracy: If $\pm\bm{x}_i^{(l)}$ is far from $\bm{y}$ in terms of Euclidean distance, then it is assumed that $\pm\bm{x}_i^{(l)}$ is not close to $\bm{y}$ in terms of distance along the class manifold. Thus $\bm{x}_i^{(l)}$ and its tangent vectors should not be needed in the $\ell^1$-minimized approximation of $\bm{y}$.

A deeper notion of the parameter $r$ is to view it as a rough estimate of the local neighborhood radius of the data set. More precisely, $r$ estimates the distance from a sample within which its class manifold can be well-approximated by a tangent hyperplane (at that sample). Given $X_{\mathrm{tr}}$ and $n$, $r$ is automatically computed, as described in Algorithm \ref{lpca_src_offline}. In words, we set $r$ to be the median distance between each training sample and its $(n+1)$st nearest neighbor (in the same class), where $n$, the number of neighbors in local PCA, is used to implicitly define the local neighborhood. It follows that $r$ is a robust estimate of the local neighborhood radius, as learned from the training data.

We verified the effectiveness of this automatically-computed parameter by comparing it to the same algorithm but with $r$ set via manual grid search during cross-validation. Though the latter method sometimes resulted in slightly higher accuracy, the saved computational expense of the automated setting of $r$ (as described above) clearly showed it to be an improvement to the overall algorithm. 

This also explains our choice for the tangent vector scaling factor $c = r \gamma$ (in Step \ref{scale_step} of Algorithm \ref{lpca_src_offline}), where $\gamma \sim \unif(0,1)$. Multiplying each tangent hyperplane basis vector $\bm{u}_j^{(l,i)}$, $1\leq j \leq d$, by this scalar and then shifting it by its corresponding training sample $\bm{x}_i^{(l)}$ helps to ensure that the resulting tangent vector, included in the dictionary $D_{\bm{y}}$ if $\pm\bm{x}_i^{(l)}$ is sufficiently close to $\bm{y}$, lies in the local neighborhood of $\bm{x}_i^{(l)}$ on the $l$th class manifold.

\begin{rem} \label{rem:r}
If the test sample $\bm{y}$ is far from the training data, defining $r$ as in Algorithm \ref{lpca_src_offline} may produce $D_{\bm{y}} = \emptyset$, i.e., there may be no training samples within that distance of $\bm{y}$. Thus to prevent this degenerate case, we use a slightly modified technique for setting $r$ in practice. After assigning the median neighborhood radius $r_1 := \mathrm{median} \big\{ r_i^{(l)}\, | \, 1\leq i \leq N_l, \, 1\leq l \leq L \big\}$, we define $r_2$ to be the distance between the test sample $\bm{y}$ and the closest training sample (up to sign). We then define the pruning parameter $r := \max\{r_1,r_2\}$. In the (degenerate) case that $r=r_2$, the dictionary consists of the closest training sample and its tangent vectors, leading to nearest neighbor classification instead of an algorithm error. However, experimental results indicate that the pruning parameter $r$ is almost always equal to the median neighborhood radius $r_1$, and so we leave this ``technicality'' out of the official algorithm statement to make it easier to interpret.
\end{rem}

\subsection{Computational Complexity and Storage Requirements}\label{sec:cc}

In this subsection, we compare the computational complexity and storage requirements of SRC and our proposed algorithm.

\subsubsection{Computational complexity of SRC}

When the $\ell^1$-minimization algorithm HOMOTOPY \cite{don:hom} is used, it is easy to see that the computational complexity of SRC is dominated by this step. This complexity is $O(N_\mathrm{tr}m\kappa+ m^2\kappa)$, where $\kappa$ is the number of HOMOTOPY iterations \cite{yan:rev}. HOMOTOPY has been shown to be relatively fast and good for use in robust face recognition \cite{yan:rev}. In our experiments, we use it in all classification methods requiring $\ell^1$-minimization.

\subsubsection{Computational complexity of LPCA-SRC}

The computational complexity of the offline phase in LPCA-SRC (Algorithm \ref{lpca_src_offline}) is
\begin{equation} \label{eq:cc_offline}
O\Big(m\sum\limits_{l=1}^L N_l^2  + N_\mathrm{tr} mn\Big),
\end{equation}
whereas that of the online phase (Algorithm \ref{lpca_src_online}) is
\begin{equation} \label{eq:cc_online}
O\Big(N_\mathrm{tr} m + \frac{N_{\bm{y}}}{d} \log\Big(\frac{N_{\bm{y}}}{d}\Big)  + N_{\bm{y}}m\kappa + m^2\kappa  \Big).
\end{equation}
Recall that $N_{\bm{y}}$ denotes the number of columns in the pruned dictionary $D_{\bm{y}}$. We note that the offline cost in Eq.~\eqref{eq:cc_offline} is based on the linear nearest neighbor search algorithm for simplicity; in practice there are faster methods. In our experiments, we used ATRIA (\emph{Advanced Triangle Inequality Algorithm} \cite{merk:knn}) via the MATLAB TSTOOL functions \texttt{nn\_prepare} and \texttt{nn\_search} \cite{merk:tstool}. The first function prepares the set of class $l$ training samples $\mathcal{X}^{(l)}$ for nearest neighbor search at the onset, with the intention that subsequent runs of \texttt{nn\_search} on this set are faster than simply doing a search without the preparation function. Other fast nearest neighbor search algorithms are available, for example, \emph{k-d tree} \cite{ben:kdtree}. The cost complexity estimates of these fast nearest neighbor search algorithms are somewhat complicated, and so we do not use them in Eq.~\eqref{eq:cc_offline}. Hence, Eq.~\eqref{eq:cc_offline} could be viewed as the worst-case scenario.

Offline and online phases combined, the very worst-case computational complexity of LPCA-SRC is $O(N_\mathrm{tr}^4)$, which occurs when the second-to-last term in Eq.~\eqref{eq:cc_online} dominates: i.e., when (i) $N_{\bm{y}} \approx (d+1) N_\mathrm{tr}$ (no pruning); (ii) $m \approx N_\mathrm{tr}$ (large relative sample dimension); (iii) very large class manifold dimension estimate $d$, so that $d$ is relatively close to $N_\mathrm{tr}$ (note that this requires very large $N_l$ for $1\leq l \leq L$ by Eq.~\eqref{eq:dn_range}, which implies that $L$ has to be very small); and (iv) $\kappa \approx m$ (many HOMOTOPY iterations). For small $\kappa$ and $N_l$, $1\leq l\leq L$, and when the pruning parameter $r$ results in small $N_{\bm{y}}$ relative to $N_\mathrm{tr}$, then the computational complexity reduces to approximately $O(N_\mathrm{tr}m)$.

\subsubsection{Storage requirements} \label{sec:storage}

The primary difference between the storage requirements for LPCA-SRC and SRC is that the offline phase of LPCA-SRC requires storing the matrix $D \in \mathbb{R}^{m \times (d+1)N_\mathrm{tr}}$, which has a factor of $d+1$ as many columns as the matrix of training samples $X_\mathrm{tr} \in \mathbb{R}^{m\times N_\mathrm{tr}}$ stored in SRC. Hence the storage requirements of LPCA-SRC are at worst $(d+1)$ times the amount of storage required by SRC.

Though this potentially is a large increase, consider that in applications such as face recognition, it is expected that the intrinsic class manifold dimension be small, e.g., 3-5 \cite{lee:linss}. Second, as we discussed in Remark \ref{rem:d} in Section \ref{sec:pars_dn}, it is often sufficient to take $d$ smaller than the actual intrinsic dimension (e.g., $d \in \{1,2\}$) in LPCA-SRC. This, combined with the assumption that the original training set in SRC is not too large (so that the $\ell^1$-minimization problem in SRC can be solved fairly efficiently), suggests that the additional storage requirements of LPCA-SRC over SRC may not deter from the use of LPCA-SRC. We discuss this further with respect to our experimental results in Section \ref{sec:exp_res}.

\section{Experiments} \label{sec:exp_res}

We tested the proposed classification algorithm on one synthetic database and three popular face\\
 databases. For all data sets, we used HOMOTOPY to solve the regularized versions of the $\ell^1$-minimization problems, i.e., Eq.~\eqref{eq:src_opt_noise} for SRC and Eq.~\eqref{eq:l1_approx} for LPCA-SRC, using version 2.0 of the L1 Homotopy toolbox \cite{asif:hom}.

\subsection{Algorithms Compared} \label{sec:algs_compared}

We compared LPCA-SRC to the original SRC, \emph{SRC}$\mathit{_{\mathrm{pruned}}}$ (a modification of SRC which we explain shortly), two versions of \emph{tangent distance classification} (our implementations are inspired by Yang et al.\ \cite{yan:ltd}), \emph{locality-sensitive dictionary learning SRC} \cite{wei:lsdl}, $\mathit{k}$\emph{-nearest neighbors classification}, and $\mathit{k}$\emph{-nearest neighbors classification over extended dictionary}. 

\begin{itemize}[noitemsep,nolistsep]

\item \textit{SRC$_\mathrm{pruned}$}: To test the efficacy of the tangent vectors in the LPCA-SRC dictionary, this modification of SRC prunes the dictionary of original training samples using the pruning parameter $r$, as in LPCA-SRC. SRC$_{\mathrm{pruned}}$ is exactly LPCA-SRC without the addition of tangent vectors.

\item \textit{Tangent distance classification} (TDC1 and TDC2): We compared LPCA-SRC to two versions of tangent distance classification to test the importance of our algorithm's sparse representation framework. Both of our implementations begin by first finding a pruned matrix $D_{\bm{y}}^{\mathrm{TDC}}$ that is very similar to the dictionary $D_{\bm{y}}$ in LPCA-SRC. In particular, $D_{\bm{y}}^{\mathrm{TDC}}$ can be found using Algorithm \ref{lpca_src_offline} and Steps 1-10 in Algorithm \ref{lpca_src_online}, \emph{omitting Step 2 in each algorithm}. That is, neither the training nor test samples are $\ell^2$-normalized in the TDC methods; compared to the SRC algorithms, TDC1 and TDC2 are not sensitive to the energy of the samples. We emphasize that the resulting matrix $D_{\bm{y}}^{\mathrm{TDC}}$ contains training samples that are nearby $\bm{y}$, as well as their corresponding tangent vectors.

In TDC1, we then divide $D_{\bm{y}}^{\mathrm{TDC}}$ into the ``subdictionaries'' 
$D_{\bm{y}}^{(l)}$, where $D_{\bm{y}}^{(l)}$ contains the portion of $D_{\bm{y}}^{\mathrm{TDC}}$ corresponding to class $l$. The test sample $\bm{y}$ is next projected onto the space spanned by the columns of $D_{\bm{y}}^{(l)}$ to produce the vector $\hat{\bm{y}}^{(l)}$, and the final classification is performed using
\begin{align*}
\classlabel(\bm{y}) = \arg \min_{1\leq l \leq L} \big\|\bm{y}-\hat{\bm{y}}^{(l)}\big\|_2.
\end{align*}

Our second implementation, TDC2, is similar. Instead of dividing $D_{\bm{y}}^{\mathrm{TDC}}$ according to class, however, we split it up according to training sample, obtaining the subdictionaries $D_{\bm{y}}^{(l,i)}$, where $D_{\bm{y}}^{(l,i)}$ contains the original training sample $\bm{x}_i^{(l)}$ and its tangent vectors. It follows that each subdictionary in TDC2 has $d+1$ columns. The given test sample $\bm{y}$ is next projected onto the space spanned by the columns of $D_{\bm{y}}^{(l,i)}$ to produce $\hat{\bm{y}}_i^{(l)}$, a vector on the (approximate) tangent hyperplane at $\bm{x}^{(l)}_i$. The final classification is performed using
\begin{align*}
\classlabel(\bm{y}) = \arg \min_{1\leq l \leq L}\Big\{\min_{1\leq i \leq N_l} \big\|\bm{y}-\hat{\bm{y}}_i^{(l)}\big\|_2\Big\}.
\end{align*}

\item \textit{Locality-sensitive dictionary learning SRC} (LSDL-SRC): Instead of directly minimizing the $\ell^1$-norm of the coefficient vector, LSDL-SRC replaces the regularization term in Eq.~\eqref{eq:src_opt_noise} of SRC with a term that forces large coefficients to occur only at dictionary elements that are close (in terms of an exponential distance function) to the given test sample. LSDL-SRC also includes a separate dictionary learning phase in which columns of the dictionary are selected from the columns of $X_\mathrm{tr}$. We note that though the name ``LSDL-SRC'' contains the term ``SRC,'' this algorithm is less related to SRC than our proposed algorithm, LPCA-SRC. See Wei et al.'s paper \cite{wei:lsdl} for their reasoning behind this name choice. However, the two algorithms do have very similar objectives, and we thought it important to compare LPCA-SRC and LSDL-SRC in order to validate our alternative approach.

\item $\mathit{k}$-\textit{nearest neighbors classification} ($k$NN): The test sample is classified to the most-represented class from among the nearest (in terms of Euclidean distance) $k$ training samples ($k$ is odd).

\item $\mathit{k}$-\textit{nearest neighbors classification over extended dictionary} ($k$NN-Ext): This is $k$NN over the columns of the (full) extended dictionary that includes the original training samples and their tangent vectors. Samples are not normalized at any stage. 

\end{itemize}

\subsection{Setting of Parameters}

For the synthetic database, we used cross-validation at each instantiation of the training set to choose the best parameters $n$, $\lambda$, and $d$ in LPCA-SRC. (Though the true class manifold dimension is known on this database, we cannot always assume that this is the case.) We optimized the parameters consecutively as described in Section \ref{sec:par_set}. We used the same approach for the parameter $\lambda$ in SRC, the parameters $n$ and $\lambda$ in SRC$_\mathrm{pruned}$, and the parameters $n$ and $d$ in the TDC algorithms. Finally, we used a similar procedure for the multiple parameters in LSDL-SRC (including its number of dictionary elements), and we also set $k$ in $k$NN and $k$NN-Ext using cross-validation.

Our approach for the face databases was very similar, though in order to save computational costs, we set some parameter values according to previously published works. In particular, we set $\lambda = 0.001$ in LPCA-SRC, SRC, and SRC$_\mathrm{pruned}$, as was used in SRC by Waqas et al.\ \cite{waq:cnrc}. Additionally, we set most of the parameters in LSDL-SRC to the values used by its authors \cite{wei:lsdl} on the same face databases, though we again used cross-validation to determine its number of dictionary elements.

\subsection{Synthetic Database} 

This subsection is organized into two parts: We describe the synthetic database in Section \ref{sec:syn_descript}, and we present our experimental findings in Section \ref{sec:syn_exps}. Figures \ref{syn_data_set_varying_class_size} and \ref{syn_data_set_varying_noise} and Table \ref{table:syn_time} show the accuracy and runtime results (as well as related information) respectively, for different versions of the synthetic database. A thorough discussion follows. Note that some algorithms from Section \ref{sec:algs_compared} (``Algorithms Compared'') have been excluded from these reported findings because of their poor performance, as we explain towards the end of Section \ref{sec:syn_exps}. Finally, we briefly discuss the storage differences between LPCA-SRC and SRC and then summarize our results on the synthetic database. 

\subsubsection{Database description} \label{sec:syn_descript}

The following synthetic database is easily visualized, and its class manifolds are nonlinear (though well-approximated by local tangent planes) with many intersections. Thus it is ideal for empirically comparing LPCA-SRC and SRC. However, we stress strongly that the classification results on this database (in Section \ref{sec:syn_exps}) are biased towards the proposed method, as the database structure is specifically designed to illustrate the advantages of LPCA-SRC over SRC. See the results on the face databases in Section \ref{sec:face} for an unbiased comparison between LPCA-SRC and the methods outlined in Section \ref{sec:algs_compared}. 

In the synthetic database, class manifolds are sinusoidal waves normalized to lie on $S^2$, with underlying equations given by
\begin{align*}
x(t) &= \cos(t+\phi), \\
y(t) &= \sin(t+\phi), \\
z(t) &= A \sin(\omega t).
\end{align*}

We set $\omega = 3$ and $A = 0.5$, and we varied $\phi$ to obtain $L$ classes. In particular, we set $\phi = 2 \pi/(3l)$ for data in class $1\leq l \leq L=4$. For each training and test set, we generated the same number $N_0 = N_l$, $l=1,\ldots,L$, of samples in each class by (i) regularly sampling $t \in [\,0,2\pi)$ to obtain the points $\bm{p}(t) = [x(t),y(t),z(t)]^\transp$; (ii) computing the normalized points $\bm{p}(t)/\|\bm{p}(t)\|_2$; (iii) appending $50$ ``noise dimensions'' to obtain vectors in $\mathbb{R}^{53}$; (iv) adding independent random noise to each coordinate of each point as drawn from the Gaussian distribution $\mathcal{N}(0,\eta^2)$; and lastly (v) re-normalizing each point to obtain vectors of length $ m = 53$ lying on $S^{m-1}$. We performed classification on the resulting data samples. Note that the reason why we turned the original $\mathbb{R}^3$ problem into a problem in $\mathbb{R}^{53}$ was because SRC is designed for high-dimensional classification problems \cite{wri:src} and to make the problem more challenging. We emphasize that we did not apply any method of dimension reduction to this database.

\begin{figure}[t]
\begin{center}
\includegraphics[width=0.3\linewidth]{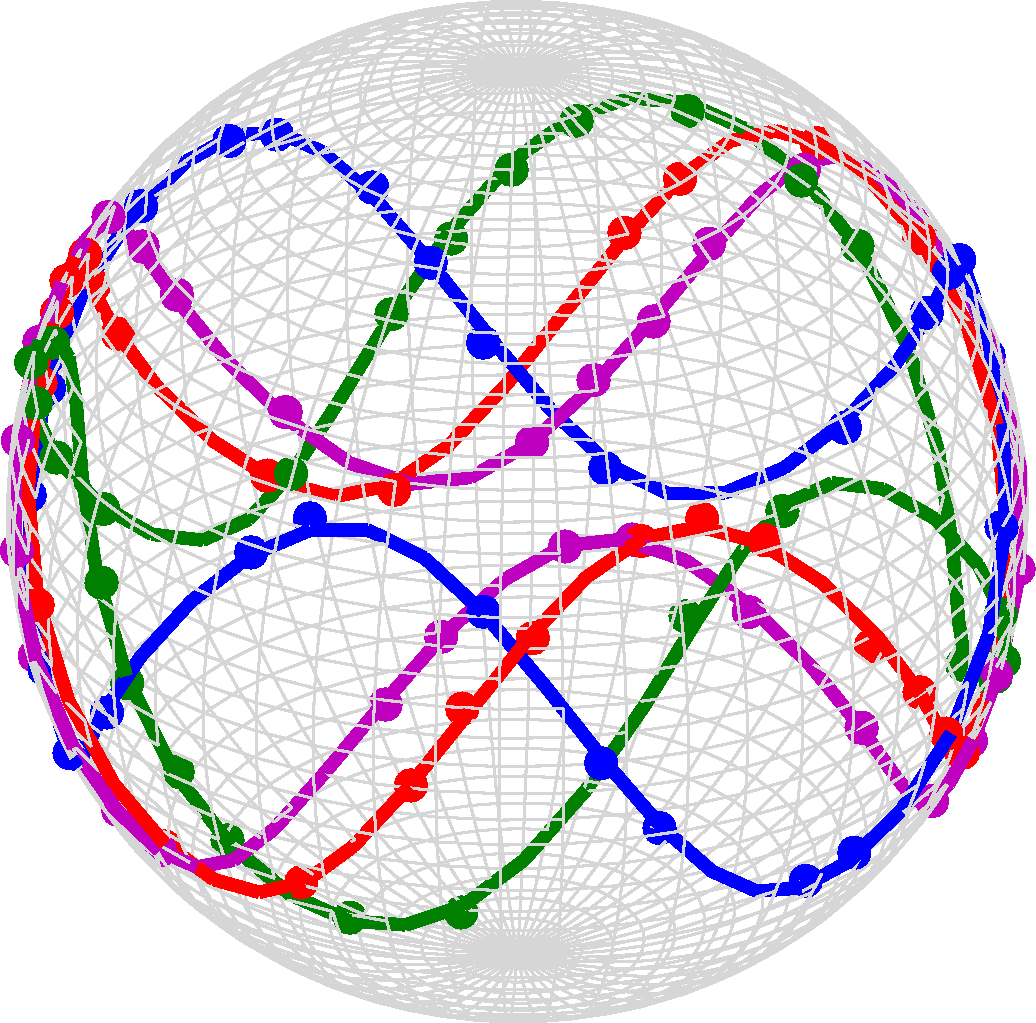}
\caption{A realization of the first three coordinates of the synthetic database training set with $N_0 = 25$ and $\eta = 0.01$. Nodes denote training samples; colors denote classes.  }
\label{syn_data_set}
\end{center}
\end{figure}

Figure \ref{syn_data_set} shows the first three coordinates of a realization of the training set of the synthetic database. Note that the class manifold dimension is the same for each class and equal to 1. The signal-to-noise ratios (SNRs) are displayed in Table \ref{tab:snr} for $N_0 = 25$ and various values of noise level $\eta$. These results were obtained by averaging the mean training sample SNR over 100 realizations of the data set.

\begin{table}[!htb] 
\centering
\begin{tabular}{|c|c|c|c|c|c|c|c|}
\hline
$\eta = 0.0001$ & $\eta = 0.001$  & $\eta = 0.005$ & $\eta = 0.01$  & $\eta = 0.015$ & $\eta = 0.02$  & $\eta = 0.03$  & $\eta = 0.05$  \\
\hline
62.85	& 42.84	&	28.86	&	22.86	&	19.35	&	16.89	&	13.45	&	9.25 \\
\hline
\end{tabular}
\caption{Mean training sample signal-to-noise ratio (in decibels) over 100 realizations of the synthetic database with $N_0 = 25$ and various values of noise level $\eta$.\label{tab:snr}}
\end{table}

\subsubsection{Experimental results} \label{sec:syn_exps}

We performed experiments on this database, first varying the number of training samples in each class and then varying the amount of noise. Table \ref{table:pars} contains brief descriptions of the relevant parameters for easy reference; a detailed description of the output parameters is given later on.

\begin{table}[!htb] 
\centering
\begin{tabular}{|c|c|c|}
\hline
Algorithm Parameters & Data Set Parameters & Output Parameters \\
\hline
$d$, $n$: Local PCA parameters & $N_0$: Class size & $N$: Dictionary size \\
$\lambda$: Error/sparsity trade-off & $\eta$: Noise level & t: Time in seconds \\
$r$: Pruning parameter (set automatically) & & $\kappa$: \# of Homotopy iterations \\
\hline
\end{tabular}
\caption{Brief descriptions of the parameters relevant to experimental results on the synthetic database.}
\label{table:pars}
\end{table}

\begin{figure}[t]
\begin{center}
\includegraphics[width=0.95\linewidth]{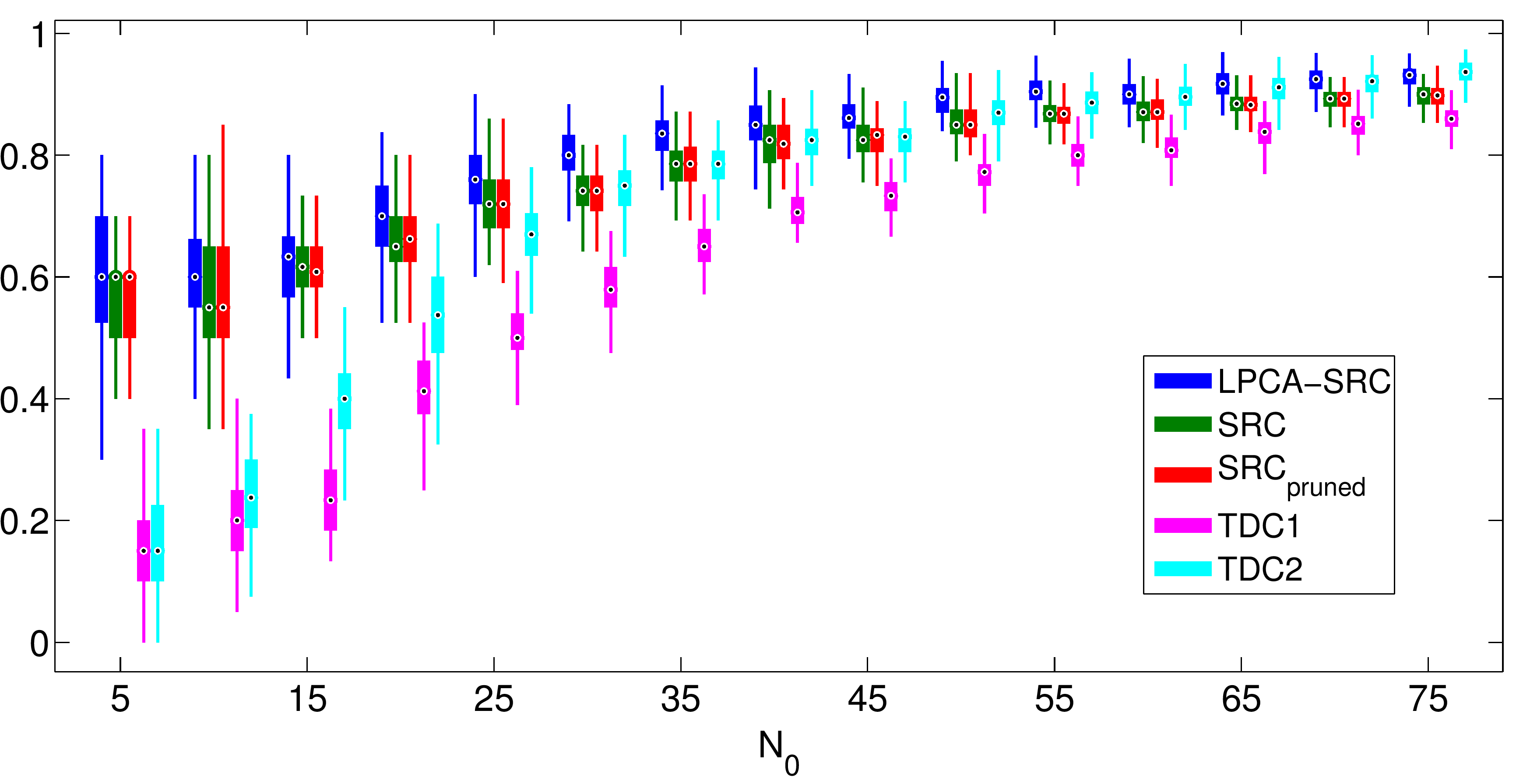}
\caption{Box plots of the average classification accuracy (over 100 trials) of competitive algorithms on the synthetic database with varying training class size $N_0$. We fixed $\eta = 0.001$.}
\label{syn_data_set_varying_class_size}
\end{center}
\end{figure}

\begin{figure}[t]
\begin{center}
\includegraphics[width=0.95\linewidth]{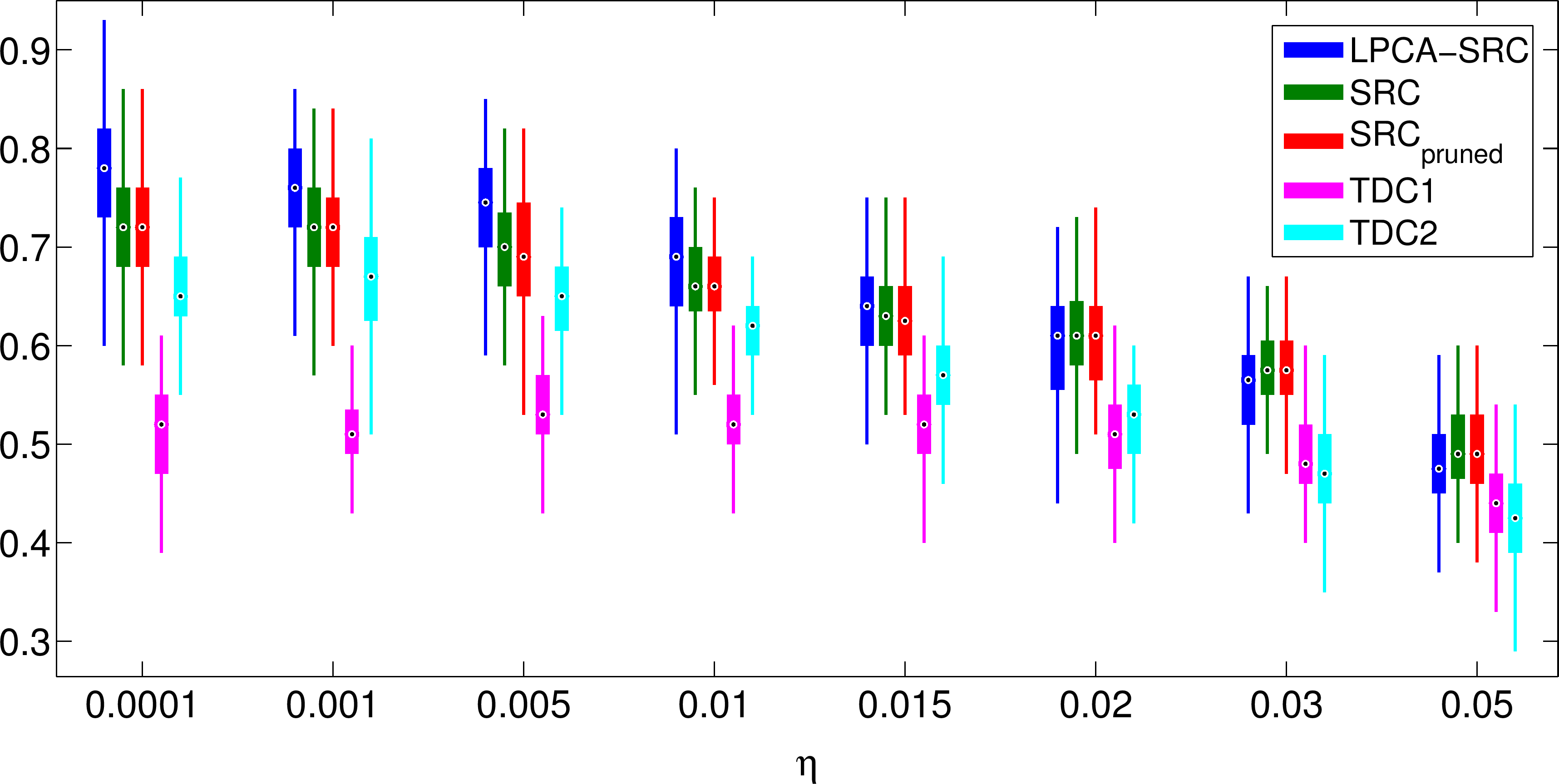}
\caption{Box plots of the average classification accuracy (over 100 trials) of competitive algorithms on the synthetic databases with varying noise level $\eta$. We fixed $N_0 = 25$.}
\label{syn_data_set_varying_noise}
\end{center}
\end{figure}

The results are presented in Figures \ref{syn_data_set_varying_class_size} and \ref{syn_data_set_varying_noise} and Table \ref{table:syn_time}; a discussion follows.

\begin{table*}[!htb] 
\small{
\centering
\begin{tabular}{|c|r|r|c|r|r|c|r|r|c|r|r|c|}
\hline
& \multicolumn{3}{|c|}{$N_0 = 5$}    & \multicolumn{3}{|c|}{$N_0 = 25$}   & \multicolumn{3}{|c|}{$N_0 = 45$}    & \multicolumn{3}{|c|}{$N_0 = 65$}   \\
\hline
Algorithm & t & $N$ & $\kappa$ & t & $N$ & $\kappa$ & t & $N$ & $\kappa$ & t & $N$ & $\kappa$ \\
\hline
LPCA-SRC	&	11.2	&	56	&	2	&	68.8	&	80	&	3	&	115.3	&	42	&	3	&	159.2	&	30	&	2	\\
SRC	      &	4.5	&	20	  &	2	&	39.9	&	100	&	3	&	104.6	&	180	&	3	&	162.8	&	260	&	3	\\
SRC$_\mathrm{\textnormal{pruned}}$	&	7.1	&	20	&	2	&	54.1	&	79	&	3	&	130.2	&	146	&	3	&	206.0	&	201	&	3	\\
TDC1	    &	10.8	&	9	&	N/A	&	43.6	&	6	&	N/A	&	71.1	&	5	&	N/A	&	92.3	&	3	&	N/A	\\
TDC2	    &	19.5	&	3	&	N/A	&	57.0	&	2	&	N/A	&	93.4	&	2	&	N/A	&	125.4	&	2	&	N/A	\\
\hline
\end{tabular}
\caption{Average runtime in ms (t), dictionary size ($N$), and number of HOMOTOPY iterations ($\kappa$) over 100 trials on the synthetic database with varying training class size $N_0$. We fixed $\eta = 0.001$.}
\label{table:syn_time}}
\end{table*}

\textbf{Accuracy results for varying class size.} Figure \ref{syn_data_set_varying_class_size} shows the average classification accuracy (over 100 trials) of the competitive algorithms as we varied the number of training samples in each class. We fixed the noise level $\eta = 0.001$. LPCA-SRC generally had the highest accuracy. On average, LPCA-SRC outperformed SRC by 3.5\%, though this advantage slightly decreased as the sampling density increased and the tangent vectors became less useful, in the sense that there were often already enough nearby training samples in the ground truth class of $\bm{y}$ to accurately approximate it without the addition of tangent vectors. SRC and SRC$_\mathrm{pruned}$ had comparable accuracy for all tried values of $N_0$, indicating that the pruning parameter $r$ was effective in removing unnecessary training samples from the SRC dictionary. Further, the increased accuracy of LPCA-SRC over SRC$_\mathrm{pruned}$ suggests that the tangent vectors in LPCA-SRC contributed meaningful class information.
 
To determine if these results are statistically significant, we performed a Repeated Measures ANOVA test on the results for LPCA-SRC, SRC, and SRC$_\mathrm{pruned}$ as well as a t-test between the results for LPCA-SRC and SRC. The detailed results can be found in Table \ref{table:anova_syn_N0} in \ref{sec:appendix}. In summary, the differences in the accuracies of LPCA-SRC, SRC, and SRC$_\mathrm{pruned}$ are statistically significant for all but $N_0 = 15$, as demonstrated by $p$-values less than $0.05$ for these experiments. 

The TDC methods performed relatively poorly for small values of $N_0$. At low sampling densities, the TDC subdictionaries were poor models of the (local) class manifolds, leading to approximations of $\bm{y}$ that were often indistinguishable from each other and resulting in poor classification. Both TDC methods improved significantly as $N_0$ increased, with TDC2 outperforming TDC1 and in fact becoming comparable to LPCA-SRC for $N_0 \geq 60$. We attribute this to the extremely local nature of TDC2: It considers a single local patch on a class manifold at a time, rather than each class as a whole. Hence under dense sampling conditions, TDC2 effectively mimicked the successful use of sparsity in LPCA-SRC.

\textbf{Accuracy results for varying noise.} Figure \ref{syn_data_set_varying_noise} shows the average classification accuracy (over 100 trials) of the competitive algorithms as we varied the amount of noise. We fixed $N_0 = 25$. LPCA-SRC had the highest classification accuracy for low values of $\eta$ (equivalently, when the SNR was high), outperforming SRC by as much as nearly $4\%$. For $\eta \geq 0.015$ (i.e., when the SNR dropped below 20 decibels), LPCA-SRC lost its advantage over SRC and SRC$_\mathrm{pruned}$. This is likely due to noise degrading the accuracy of the tangent vectors. SRC and SRC$_\mathrm{pruned}$ had nearly identical accuracy for all values of $\eta$; again, this illustrates that faraway training samples (as defined by the pruning parameter $r$) did not contribute to the $\ell^1$-minimized approximation of the test sample, and the increased accuracy of LPCA-SRC over SRC$_\mathrm{pruned}$ for low noise values demonstrates the efficacy of the tangent vectors in LPCA-SRC in these cases. We briefly note that when we vary the noise level for larger values of $N_0$, the accuracy of the tangent vectors generally improves. As a result, we see that LPCA-SRC can tolerate higher values of $\eta$ before being outperformed by SRC and SRC$_\mathrm{pruned}$. 

Table \ref{table:anova_syn_eta} in \ref{sec:appendix} contains the $p$-values for rANOVA and related tests on the accuracy results of LPCA-SRC, SRC, and SRC$_\mathrm{pruned}$, as well as the 5\% confidence intervals for the advantage of LPCA-SRC over SRC. These tests concur with the discussion above; LPCA-SRC outperforms SRC for small values of $\eta$, there is no clear advantage for $\eta \in \{0.01, 0.015\}$, and SRC outperforms LPCA-SRC for $\eta \geq 0.02$.

TDC2 outperformed TDC1 for all but the largest values of $\eta$, though both algorithms were outperformed by the three SRC methods at this relatively low sampling density for the reasons discussed previously. For $\eta \geq 0.03$, TDC2 began performing worse than TDC1. We expect that the local patches represented by the subdictionaries in TDC2 became poor estimates of the (tangent hyperplanes of the) class manifolds as the noise increased, resulting in a decrease in classification accuracy.

\textbf{Runtime results for varying class size.} In Table \ref{table:syn_time}, we display the runtime-related information of the competitive algorithms with varying training class size. (We do not show the runtime results for the case of varying noise; the results for varying class size are much more revealing.) In particular, we report the average runtime (in milliseconds), the number of columns in each algorithm's dictionary (we refer to this as the ``size'' of the dictionary, as the sample dimension is fixed), and the number of HOMOTOPY iterations. These latter variables are denoted $N$ and $\kappa$, respectively. The runtime does not include the time it took to perform cross-validation and is the total time (averaged over 100 trials) of performing classification on the entire database. In the case that the algorithm has separate offline and online phases (e.g., LPCA-SRC), both phases are included in this total. For the TDC methods, we report the average subdictionary sizes,\footnote{Recall that these subdictionaries are the class-specific portions $D_{\bm{y}}^{(l)}$, $1\leq l \leq L$, of the main dictionary $D_{\bm{y}}^{\mathrm{TDC}}$. Thus the values of $N$ for TDC1 and TDC2 are much smaller than those for the other classification methods.} and for conciseness, we display the results for only a handful of the values of $N_0$. We use ``N/A'' to indicate that a particular statistic is not applicable to the given algorithm.

The dictionary sizes of LPCA-SRC, SRC, and SRC$_\mathrm{\textnormal{pruned}}$ are quite informative. Recall that LPCA-SRC outperformed SRC and SRC$_\mathrm{pruned}$ (by more than 3\%) for the shown values of $N_0$. For $N_0 = 5$, the dictionary in LPCA-SRC was larger than that of the two other methods, adaptively retaining more samples to counter-balance the low sampling density. At large values of $N_0$, LPCA-SRC took full advantage of the increased sampling density, stringently pruning the set of training samples and keeping only those very close to $\bm{y}$. Due to the resulting small dictionary, it had comparable runtime to SRC despite its additional cost of computing tangent vectors. In contrast, without the addition of tangent vectors, SRC$_\mathrm{pruned}$ was forced to keep a large number of training samples in its dictionary; the cost of the dictionary pruning step resulted in SRC$_\mathrm{pruned}$ running slower than SRC, despite its slightly smaller dictionary. (We note that one might expect that SRC$_\mathrm{pruned}$ would always have a smaller dictionary than LPCA-SRC since it does not include tangent vectors; this is not the case, as the value of the number-of-neighbors parameter $n$, and hence the pruning parameter $r$, may be different for the two algorithms.)

The TDC methods ran relatively fast, especially for large values of $N_0$. This is expected, as these algorithms do not require $\ell^1$-minimization.

\textbf{Summary.} The experimental results on the synthetic database show that LPCA-SRC can achieve higher classification accuracy than SRC and similar methods when the class manifolds are sparsely sampled and the SNR is large. In these cases, the tangent vectors in LPCA-SRC help to ``fill out'' portions of the class manifolds that lack training samples. When the sampling density was sufficiently high, however, we saw that the tangent vectors in LPCA-SRC were less needed to provide an accurate, local approximation of the test sample, and thus LPCA-SRC offered a smaller advantage over SRC and SRC$_\mathrm{pruned}$. Additionally, for higher noise (i.e., low SNR) cases, the computed tangent vectors were less reliable and the classification performance consequently deteriorated. With regard to runtime, LPCA-SRC appeared to adapt to the sampling density of the synthetic database, and though the addition of tangent vectors initially increased the dictionary size in LPCA-SRC, the online dictionary pruning step allowed for runtime comparable to SRC when the class sizes were large. 

\subsection{Face Databases} \label{sec:face}

This subsection is organized as follows: 
\begin{itemize} [noitemsep,nolistsep]
\item We first explain our experimental setup. We describe the different face databases and state the training set sizes in Section \ref{sec:face_descript}, and in Sections \ref{sec:dr} and \ref{sec:occ}, we describe the method of dimension reduction used on the raw samples and our approach to handling data samples with occlusion, respectively. Section \ref{sec:tab_pars} simply contains Table \ref{table:pars_face}, which shows brief descriptions of the relevant parameters on the face databases for easy reference.

\item We separate our classification results into two parts: Section \ref{sec:AR} contains our results on the AR face database, and Section \ref{sec:yale_orl} contains our results on the Extended Yale B and ORL face databases. More precisely, Figures \ref{AR-1_results_acc}-\ref{AR-2_results_acc} and Table \ref{AR_results_time} contain the accuracy and runtime results for two versions of the AR face database; Figures \ref{Ext-Yale-B_results_acc}-\ref{ORL_results_acc} and Tables \ref{Ext_Yale_B_time}-\ref{ORL_time} show the same results for Extended Yale B and ORL. Again, these databases are described in Section \ref{sec:face_descript}. The figures and tables in each section are followed by a discussion of their results. 

\item In Section \ref{sec:pca_tv}, we offer evidence to support our claim that the tangent vectors in LPCA-SRC can recover discriminative information lost during PCA transforms to low dimensions. We display the PCA-recovered tangent vectors and compare them to the original samples (without PCA transform) as well as the recovered samples (after PCA transform). 

\item Lastly, Section \ref{sec:face_sum} contains a summary of our experimental findings on the face databases. 
\end{itemize}

\subsubsection{Database description} \label{sec:face_descript}

The \emph{AR Face Database} \cite{AR:face} contains 70 male and 56 female subjects photographed in two separate sessions held on different days. Each session produced 13 images of each subject, the first seven with varying lighting conditions and expressions, and the remaining six images occluded by either sunglasses or scarves under varying lighting conditions. Images were cropped to $165 \times 120$ pixels and converted to grayscale. In our experiments, we selected the first 50 male subjects and first 50 female subjects, as was done in several papers (e.g., Wright et al.\ \cite{wri:src}), for a total of 100 classes. We performed classification on two versions of this database. The first, which we call ``AR-1,'' contains the 1400 un-occluded images from both sessions. The second version, ``AR-2,'' consists of the images in AR-1 as well as the 600 occluded images (sunglasses and scarves) from Session 1.

The \emph{Extended Yale Face Database B} \cite{geo:illum} contains $38$ classes (subjects) with about $64$ images per class. The subjects were photographed from the front under various lighting conditions. We used the version of Extended Yale B that contains manually-aligned, cropped, and resized images of dimension $192 \times 168$.

The \emph{Database of Faces} (formerly ``The ORL Database of Faces'') \cite{att:orl} contains $40$ classes (subjects) with $10$ images per class. The subjects were photographed from the front against dark, homogeneous backgrounds. The sets of images of some subjects contain varying lighting conditions, expressions, and facial details. Each image in ORL is initially of $92 \times 112$ pixels.

Given existing work on the manifold structure of face databases (e.g., that of Saul and Roweis \cite{row:lle}, He et al.\ \cite{he:lapface}, and Lee et al.\ \cite{lee:linss}), we make the following suppositions: Since images in each class in AR-1 and AR-2 have extreme variations in lighting conditions and differing expressions, the class manifolds of these databases may be nonlinear. Further, the natural occlusions contained in AR-2 make these class manifolds \emph{highly} nonlinear. Alternatively, since the images in each class in Extended Yale B differ primarily in lighting conditions, the class manifolds may be nearly linear. Lastly, since the images in some classes in ORL differ in both lighting conditions and expression, these class manifolds may be nonlinear; however, since the variations are small, these manifolds may be well-approximated by linear subspaces.

With regard to sampling density, we reiterate that Extended Yale B has large class sizes compared to AR and ORL. In our experiments, we randomly selected the same number of samples in each class to use for training, i.e., we set $N_0 \equiv N_l$, $1\leq l \leq L$, where $N_0$ was half the number of samples in each class.\footnote{Since the class sizes vary slightly in Extended Yale B, we set $N_0 = 32$ on this database.} We used the remaining samples for testing.

\subsubsection{Dimension reduction} \label{sec:dr}
 
To perform dimension reduction on the face databases, we used (global) PCA to transform the raw images to $m_{\mathrm{PCA}} \in \{30, 56,120\}$ dimensions before performing classification. Similar values for $m_{\mathrm{PCA}}$ were used by Wright et al.\ \cite{wri:src}. For the remainder of this paper, we will refer to the PCA-compressed versions of the raw face images as ``feature vectors'' and $m_\mathrm{PCA}$ as the ``feature dimension.'' We note that the data was not centered (around the origin) in the PCA transform space.

\subsubsection{Handling occlusion} \label{sec:occ}

Since AR-2 contains images with occlusion, we considered using the ``occlusion version'' of SRC (with analogous modifications to LPCA-SRC and SRC$_\mathrm{pruned}$) on this database. As discussed by Wright et al.\ \cite{wri:src}, this model assumes that $\bm{y}$ is the summation of the (unknown) true test sample $\bm{y}_0$ and an (unknown) sparse error vector. The resulting modified $\ell^1$-minimization problem consists of appending the dictionary of training samples with the identity matrix $I \in \mathbb{R}^{m\times m}$ and decomposing $\bm{y}$ over this augmented dictionary. For more details, see Section 3.2 of the SRC paper \cite{wri:src}.

However, the context in which Wright et al.\ use the occlusion version of SRC on the AR database is critically different than our experimental setup here \cite{wri:src}. In the SRC paper, the samples with occlusion make up the test set. In our case, both the training and test set contain samples with and without occlusion. As a consequence, occluded samples in the training set can be used to express test samples with occlusion, and on the other hand, the use of the identity matrix to extend the dictionary in SRC results in too much error allowed in the approximation of un-occluded samples. Correspondingly, we see much worse classification performance in SRC when we use its occlusion version on AR-2. Hence, we stick to Algorithm \ref{alg:src} (the original version of SRC) on all face databases.

\subsubsection{Table of Parameters} \label{sec:tab_pars}

Table \ref{table:pars_face} contains brief descriptions of the parameters relevant to the face databases.

\begin{table}[!htb] 
\centering
\begin{tabular}{|c|c|c|}
\hline
Algorithm Parameters & Data Set Parameters & Output Parameters \\
\hline
$d$, $n$: Local PCA parameters & $N_0$: Class size & $N$: Dictionary size \\
$\lambda$: Error/sparsity trade-off & $m_\mathrm{PCA}$: PCA dimension & t: Time in seconds \\
$r$: Pruning parameter (set automatically) & & $\kappa$: \# of Homotopy iterations \\
\hline
\end{tabular}
\caption{Brief descriptions of the parameters relevant to experimental results on the face databases.}
\label{table:pars_face}
\end{table}

\subsubsection{AR Face Database results} \label{sec:AR}


\begin{figure}[t]
\begin{center}
\includegraphics[width=\linewidth]{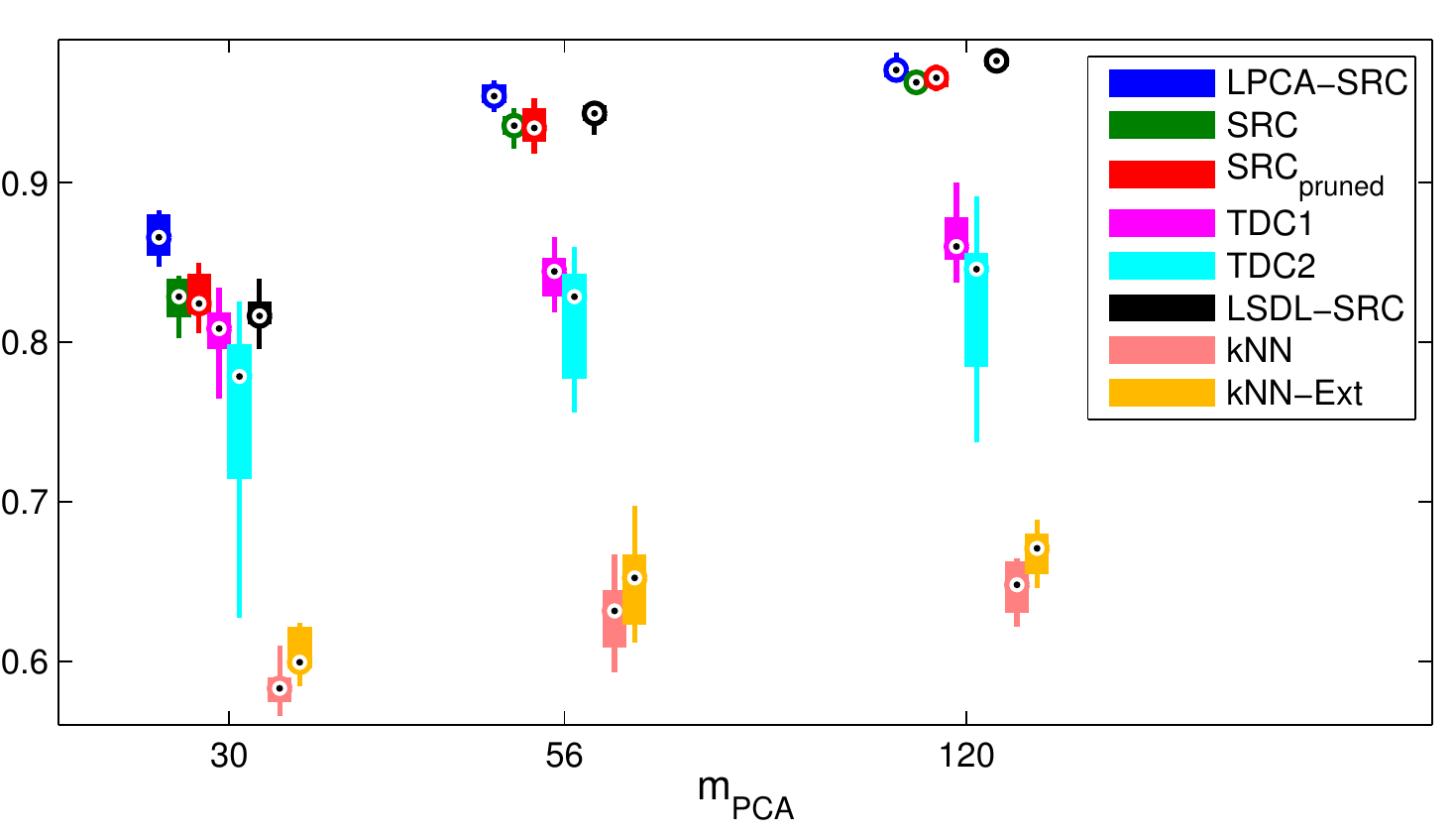}
\caption{Box plots of the average classification accuracy (over 10 trials) on the AR-1 face database for different values of $m_\mathrm{PCA}$.}
\label{AR-1_results_acc}
\end{center}
\end{figure}

\begin{figure}[t]
\begin{center}
\includegraphics[width=\linewidth]{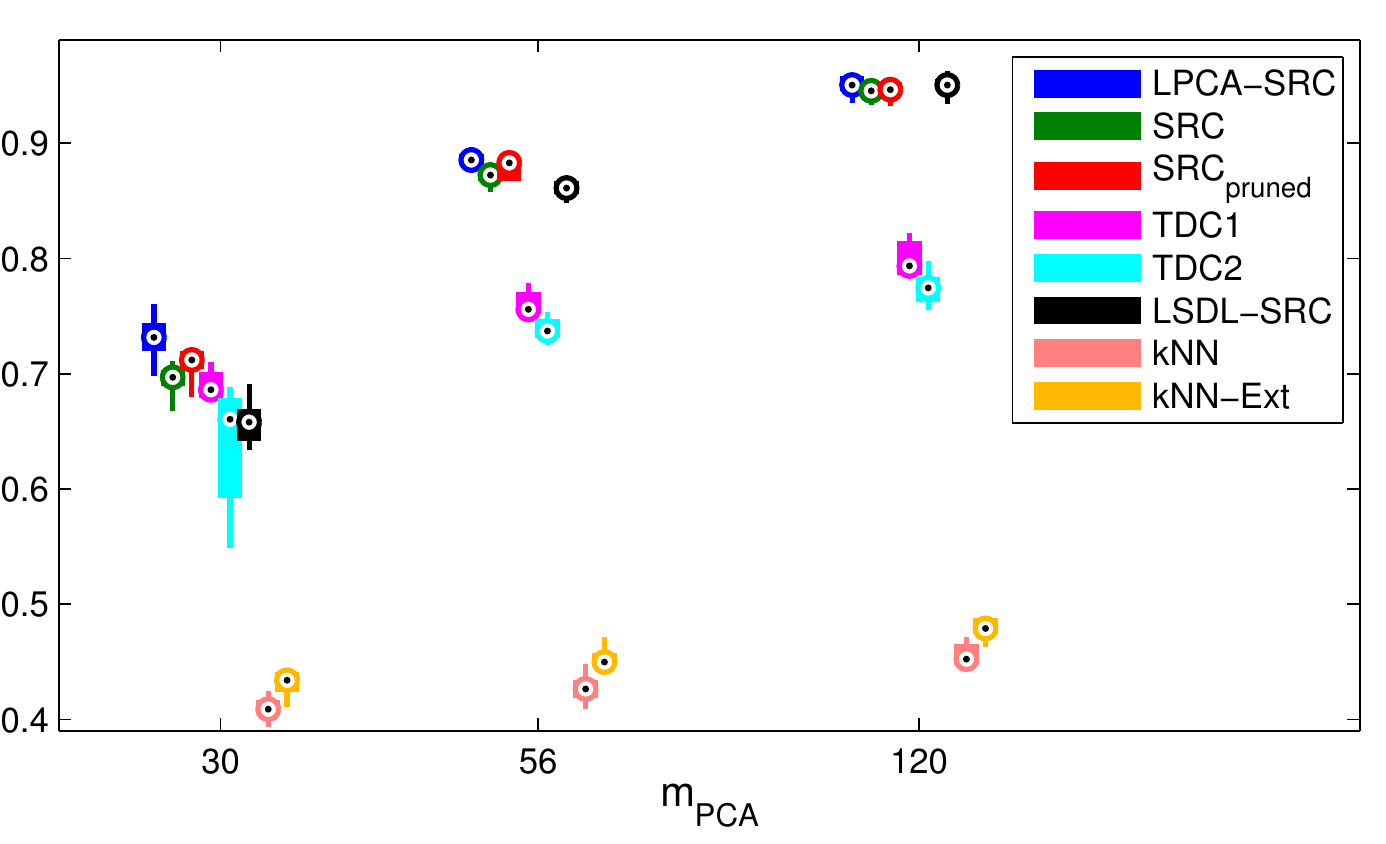}
\caption{Box plots of the average classification accuracy (over 10 trials) on the AR-2 face database for different values of $m_\mathrm{PCA}$.}
\label{AR-2_results_acc}
\end{center}
\end{figure}

\begin{table*}[!htb] 
\footnotesize{
\centering
\begin{tabular}{|c|r|r|c|r|r|c|r|r|c|}
\hline
& \multicolumn{9}{|c|}{AR-1}\\
\hline
& \multicolumn{3}{|c|}{$m_\mathrm{PCA} = 30$} &  \multicolumn{3}{|c|}{$m_\mathrm{PCA} = 56$}    & \multicolumn{3}{|c|}{$m_\mathrm{PCA} = 120$}\\
\hline
Algorithm & t & $N$ & $\kappa$ & t & $N$ & $\kappa$ & t & $N$ & $\kappa$ \\
\hline
LPCA-SRC	          & 7253	& 435	& 61	& 12496	& 676	& 87	 & 19068	& 795	& 112	\\
SRC	                & 6114	& 700	& 51	& 8875	& 700	& 72   & 13574	& 700	& 99\\
SRC$_\mathrm{pruned}$& 3763	& 231	& 39	& 5099	& 226	& 49	 & 6897	& 232	& 60 \\
TDC1	              &	11816	&	16	&	N/A	&	14239	&	16	&	N/A  &	24296	&	19	&	N/A	\\
TDC2	              &	8895	&	5	  &	N/A	&	16786	&	5	  &	N/A  &	36682	&	5	  &	N/A	 \\
LSDL-SRC	          & 7776	& 440	& N/A & 8552	& 470	& N/A  & 9720	& 490	& N/A \\
$k$NN               & 13	  & 700 & N/A	&	18	  & 700 & N/A	 &	29	& 700 & N/A \\
$k$NN-Ext           & 102	  & 2170& N/A	&	132	  & 2240 & N/A &	253	& 2660 & N/A \\

\hline
& \multicolumn{9}{|c|}{AR-2}\\
\hline
& \multicolumn{3}{|c|}{$m_\mathrm{PCA} = 30$} &  \multicolumn{3}{|c|}{$m_\mathrm{PCA} = 56$}    & \multicolumn{3}{|c|}{$m_\mathrm{PCA} = 120$}\\
\hline
Algorithm & t & $N$ & $\kappa$ & t & $N$ & $\kappa$ & t & $N$ & $\kappa$ \\
\hline
LPCA-SRC	          & 10533	 & 478	& 58   & 35269	 & 1593	  & 10     & 56169	  & 1690	& 151 \\
SRC	                & 11394	 & 1000	& 58   & 17674	 & 1000	  & 85     & 27743  	& 1000	& 121 \\
SRC$_\mathrm{pruned}$& 11118	 & 788	& 54 & 16631	 & 775	  & 77     & 24880	  & 767	  & 107 \\
TDC1	              &	20557	 &	25	&	N/A  &	27515	 &	26	  &	N/A    &	43073	  &	26	  &	N/A \\
TDC2	              &	20930	 &	6	  &	N/A  &	47571	 &	6	    &	N/A    &	103796	&	6	    &	N/A \\
LSDL-SRC	          & 22698	 & 750	& N/A  & 16337	 & 620	  & N/A    & 22191	  & 710	  & N/A \\
$k$NN               &	 15  & 1000  & N/A	 &	21	 & 1000    & N/A 	   &	37	  & 1000   & N/A \\
$k$NN-Ext           & 128	 & 4300  & N/A	 &	152	 & 3600    & N/A     &	294	  & 4400   & N/A \\

\hline
\end{tabular}
\caption{Average runtime in ms (t), dictionary size ($N$), and number of HOMOTOPY iterations ($\kappa$) over 10 trials on AR.} 
\label{AR_results_time}}
\end{table*}

\textbf{Accuracy results on AR.} Figures \ref{AR-1_results_acc} and \ref{AR-2_results_acc} display the accuracy results over 10 trials for the two versions of AR, respectively. LPCA-SRC had substantially higher classification accuracy than the other methods on both versions of AR with $m_\mathrm{PCA} = 30$. This suggests that the tangent vectors in LPCA-SRC were able to recover important class information lost in the stringent PCA dimension reduction. As $m_\mathrm{PCA}$ increased, however, the methods SRC, SRC$_\mathrm{pruned}$, and LSDL-SRC became more competitive, as more discriminative information was retained in the feature vectors and less needed to be provided by the LPCA-SRC tangent vectors. SRC$_\mathrm{pruned}$ had comparable accuracy to SRC, indicating that, once again, training samples could be removed from the SRC dictionary using the pruning parameter $r$ without decreasing classification accuracy. In some cases, the removal of these faraway training samples slightly improved class discrimination.

To test for statistical significance in the differences between accuracy results, we performed a Repeated Measures ANOVA test on LPCA-SRC, SRC, SRC$_\mathrm{pruned}$, and LSDL-SRC as well as two t-tests, one between LPCA-SRC and SRC and the other between LPCA-SRC and LSDL-SRC. The related $p$-values and confidence intervals are contained in Table \ref{table:anova_face_AR} in \ref{sec:appendix}. In summary, LPCA-SRC outperforms both methods in a statistically-significant manner, except for LSDL-SRC when $m_\mathrm{PCA} = 120$. 

For the most part, the other algorithms performed poorly on AR. The exception was LSDL-SRC, which had comparable accuracy to LPCA-SRC for $m_\mathrm{PCA} = 120$ (slightly outperforming it for AR-1) and beat SRC on AR-1 for $m_\mathrm{PCA} = 56$. However, LSDL-SRC had lower accuracy than the SRC algorithms for $m_\mathrm{PCA} = 30$ on both versions of this database. In contrast, the TDC methods performed relatively better for $m_\mathrm{PCA}=30$ than for larger values of $m_\mathrm{PCA}$ due to their more effective use of tangent vectors at this small feature dimension. Overall, however, their class-specific dictionaries were not as effective on this nonlinear, sparsely sampled database as the multi-class dictionaries of the previously-discussed algorithms. Further, TDC2 often had notably high standard error, presumably because of its sensitivity to the value of the manifold dimension estimate $d$. This could perhaps be mitigated by using a different cross-validation procedure. Lastly, $k$NN and $k$NN-Ext had the lowest classification accuracies, though $k$NN-Ext offered a slight improvement over $k$NN. Both methods consistently selected $k=1$ during cross-validation.

\textbf{Runtime results on AR.} Table \ref{AR_results_time} displays the average runtime and related results (over 10 trials) of the various classification algorithms for both versions of AR. Again, the runtime does not include the time it took to perform cross-validation and is the total time (averaged over 10 trials) of performing classification on the entire database (offline and online phases both included when applicable). The ``dictionary size'' $N$ for $k$NN and $k$NN-Ext refers to the average size of the set from which the $k$-nearest neighbors are selected (e.g., for $k$NN, $N = N_\text{tr}$). 

The generally large dictionary sizes of LPCA-SRC (and its consequently long runtimes) indicate that minimal dictionary pruning often occurred. Thus LPCA-SRC was generally slower than SRC and SRC$_\mathrm{pruned}$. However, on AR-2 with $m_\mathrm{PCA}=30$, LPCA-SRC was able to eliminate many training samples from its dictionary, due to its effective use of tangent vectors on the (presumably) highly-nonlinear class manifolds of AR-2. At this low feature dimension, the computed tangent vectors contained more class discriminative information than nonlocal training samples, likely allowing for a more accurate---and local---approximation of $\bm{y}$ on its ground truth class manifold. LPCA-SRC was faster than SRC and SRC$_\mathrm{pruned}$ (which kept a large number of training samples) in this case, and this is impressive, considering that LPCA-SRC also outperformed these methods by nearly $4\%$ and more than $2\%$, respectively.

Despite not requiring $\ell^1$-minimization, the TDC methods were often the slowest algorithms on the AR databases. We suspect that this is largely due to the relatively large number of classes in AR---recall that both TDC methods must compute least squares solutions (in TDC2, sometimes many of them) for each class represented in the pruned dictionary $D_{\bm{y}}^{\mathrm{TDC}}$. Further, TDC2 selected a relatively large value of $d$ during cross-validation (presumably so that its subdictionaries would contain a wider ``snapshot'' of the class manifolds), which made it even less efficient. The runtime of LSDL-SRC, unlike those of most of the other algorithms, was fairly insensitive to the feature dimension, and as a result, LSDL-SRC was relatively efficient for $m_\mathrm{PCA} \in \{56,120\}$. However, the expense of its dictionary learning phase for $m_\mathrm{PCA}=30$, at which the $\ell^1$-minimization algorithm in the SRC methods could be solved efficiently, resulted in LSDL-SRC's relatively slow runtime. Both $k$NN methods ran significantly faster than all the other methods.

\subsubsection{Extended Yale Face Database B and Database of Faces (``ORL'') results} 
\label{sec:yale_orl}

\begin{figure}[t]
\begin{center}
\includegraphics[width=\linewidth]{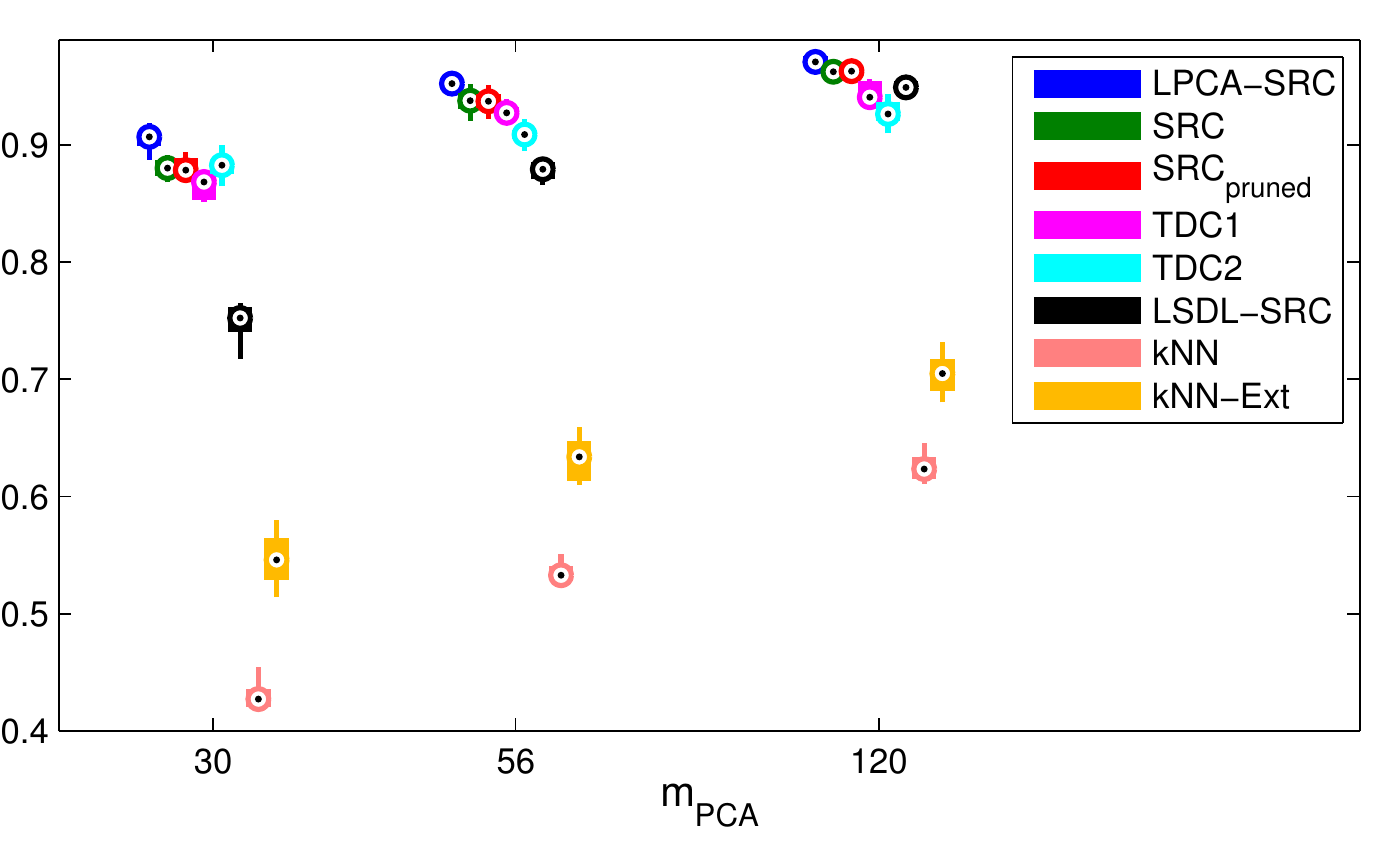}
\caption{Box plots of the average classification accuracy (over 10 trials) on the Extended Yale B face database for different values of $m_\mathrm{PCA}$.}
\label{Ext-Yale-B_results_acc}
\end{center}
\end{figure}

\begin{figure}[t]
\begin{center}
\includegraphics[width=\linewidth]{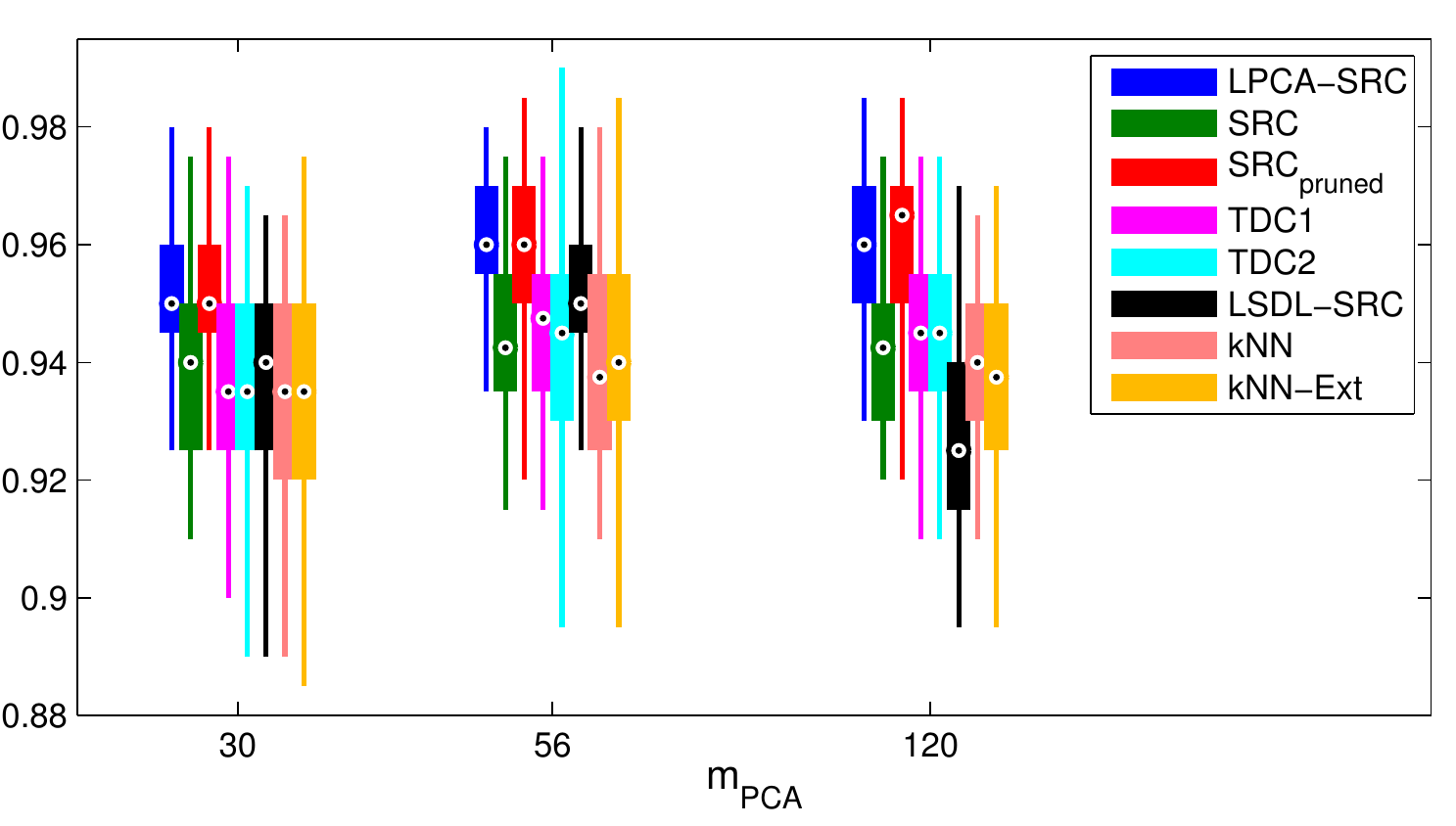}
\caption{Box plots of the average classification accuracy (over 50 trials) on the ORL face database for different values of $m_\mathrm{PCA}$.}
\label{ORL_results_acc}
\end{center}
\end{figure}

\begin{table*}[!htb] 
\small{
\centering
\begin{tabular}{|c|r|r|c|r|r|c|r|r|c|}
\hline
& \multicolumn{3}{|c|}{$m_{\mathrm{PCA}} = 30$} & \multicolumn{3}{|c|}{$m_{\mathrm{PCA}} = 56$} & \multicolumn{3}{|c|}{$m_{\mathrm{PCA}} = 120$} \\
\hline
Algorithm & t & $N$ & $\kappa$ & t & $N$ & $\kappa$ & t & $N$ & $\kappa$ \\
\hline
LPCA-SRC	& 29204	& 1922	& 75	& 72122	& 3359	& 120	& 141966	& 3785	& 182 \\
SRC	& 15584	& 1216	& 62	& 24697	& 1216	& 91	& 41939	& 1216	& 137 \\
SRC$_\mathrm{pruned}$ & 15915	& 1111	& 61	& 23813	& 1112	& 88	& 40504	& 1115	& 131 \\
TDC1 &	8098	&	20	&	N/A	&	27620	&	59	&	N/A	&	42828	&	59	&	N/A	\\
TDC2 &	11675	&	6	&	N/A	&	23506	&	6	&	N/A	&	56006	&	6	&	N/A	\\
LSDL-SRC	& 67295	& 1186	& N/A	& 53031	& 1003	& N/A	& 38731	& 821	& N/A \\
$k$NN &	17	& 1216 & N/A &	26	& 1216 & N/A &	49	& 1216 & N/A \\
$k$NN-Ext &	172	& 5350 & N/A &	251	& 4742 & N/A &	443	& 4864 & N/A \\	
\hline
\end{tabular}
\caption{Average runtime in ms (t), dictionary size ($N$), and number of HOMOTOPY iterations ($\kappa$) over 10 trials on Extended Yale B.} 
\label{Ext_Yale_B_time}}
\end{table*}

\begin{table*}[!htb] 
\small{
\centering
\begin{tabular}{|c|r|r|c|r|r|c|r|r|c|}
\hline
& \multicolumn{3}{|c|}{$m_{\mathrm{PCA}} = 30$} & \multicolumn{3}{|c|}{$m_{\mathrm{PCA}} = 56$} & \multicolumn{3}{|c|}{$m_{\mathrm{PCA}} = 120$} \\
\hline
Algorithm & t & $N$ & $\kappa$ & t & $N$ & $\kappa$ & t & $N$ & $\kappa$ \\
\hline
	LPCA-SRC	&	539	&	59	&	26	&	730	&	72	&	34	&	1221	&	111	&	50	\\
	SRC	&	854	&	200	&	40	&	1337	&	200	&	57	&	2087	&	200	&	81	\\
	SRC$_\mathrm{pruned}$	&	254	&	19	&	12	&	343	&	26	&	16	&	530	&	39	&	24	\\
	TDC1	&	121	&	1	&	N/A	&	162	&	1	&	N/A	&	344	&	1	&	N/A	\\
	TDC2	&	117	&	3	&	N/A	&	233	&	3	&	N/A	&	532	&	3	&	N/A	\\
	LSDL-SRC	&	1040	&	116	&	N/A	&	1088	&	121	&	N/A	&	931	&	102	&	N/A	\\
	$k$NN	&	8	&	200	&	N/A	&	8	&	200	&	N/A	&	9	&	200	&	N/A	\\
	$k$NN-Ext	&	25	&	568	&	N/A		&	28	&	592	&	N/A		&	38	&	568	&	N/A	\\

\hline
\end{tabular}
\caption{Average runtime in ms (t), dictionary size ($N$), and number of HOMOTOPY iterations ($\kappa$) over 50 trials on ORL.} 
\label{ORL_time}}
\end{table*}

\textbf{Accuracy results on Extended Yale B and ORL.} Figure \ref{Ext-Yale-B_results_acc} displays the accuracy results for Extended Yale B (over 10 trials), and Figure \ref{ORL_results_acc} displays the accuracy results for ORL (over 50 trials). On Extended Yale B, LPCA-SRC had the highest accuracy for all $m_\mathrm{PCA}$, though as we saw on the AR database, this advantage decreased as $m_\mathrm{PCA}$ increased and SRC became more competitive. SRC and SRC$_\mathrm{pruned}$ had similar accuracy, indicating that training samples excluded from the dictionary via the pruning parameter $r$ did not provide class information. TDC1 and TDC2 had consistently mediocre performance, neither one outperforming the other over all settings of $m_\mathrm{PCA}$, and LSDL-SRC improved as $m_\mathrm{PCA}$ increased, analogous to its behavior on AR. However, LSDL-SRC was outperformed by LPCA-SRC, even for $m_\mathrm{PCA}=120$, suggesting that the improved approximations in LPCA-SRC via its use of tangent vectors were more effective (even at this high feature dimension) than the procedure in LSDL-SRC. Along these same lines, the tangent vectors in $k$NN-Ext offered a considerable improvement over $k$NN, though once again both methods reported lower accuracy than all the other algorithms. As on AR, the $k$NN methods consistently selected $k=1$ during cross-validation.

On ORL, LPCA-SRC and SRC$_\mathrm{pruned}$ had comparable accuracy and outperformed SRC. This indicates that: (i) the pruning parameter $r$ in LPCA-SRC and SRC$_\mathrm{pruned}$ was \emph{helpful} to classification (instead of simply being benign); and (ii) the tangent vectors computed in LPCA-SRC were not. With regard to (i), it must be the case that faraway training samples---those in different classes from the test sample---contributed significantly to the approximation of the test sample in SRC, negatively affecting classification performance. This is an example of \emph{sparsity not necessarily leading to locality} (as it is relevant to class discrimination), as discussed in the LSDL-SRC paper \cite{wei:lsdl}. With regard to (ii), we suspect that the tangent vectors in LPCA-SRC were simply \emph{unneeded} to improve the classification performance on ORL. Though the approximations in SRC contained nonzero coefficients at training samples not in the same class as $\bm{y}$---presumably because of the sparse sampling and nonlinear structure of the class manifolds---many of these wrong-class training samples could be eliminated simply based on their distance to $\bm{y}$. This suggests that ORL's class manifolds can be fairly well-separated via Euclidean distance. An additional reason for (ii) was because the PCA transform to the dimensions specified in this experiment did not result in a loss of too much information, at least compared to AR and Extended Yale B. See Table \ref{tab:energy} at the end of Section \ref{sec:pca_tv} for this comparison.

As we did for the AR face database, we performed statistical analysis on the reported accuracies for Extended Yale B and ORL. The detailed results are contained in Table \ref{table:anova_face_yale_orl} in \ref{sec:appendix}. In summary, LPCA-SRC outperforms both SRC and LSDL-SRC with 95\% confidence in all of these experiments, albeit its lift in accuracy is sometimes small.

All of the remaining methods performed relatively well on ORL. The accuracies of TDC1 and TDC2 were similar and comparable to those of SRC. We ascertained that the success of the TDC methods was not due to their use of tangent vectors but instead the result of their ``per-class'' approximations of the test sample. This approach was very effective on the (presumably) well-separated class manifolds of ORL. Strikingly, the accuracy of LSDL-SRC was relatively low for $m_\mathrm{PCA} = 120$, opposite to the trend we saw on the previous face databases. The performance of LSDL-SRC could be improved for $m_\mathrm{PCA}=120$ on this database if the samples were centered (around the origin) after PCA dimension reduction. However, we confirmed that LDSL-SRC was still outperformed by LPCA-SRC in this case (albeit by a smaller margin), and its performance with centering on the other face databases was much worse than our reported results. In contrast to the results on Extended Yale B, $k$NN-Ext only provided a slight increase in accuracy over $k$NN, with the tangent vectors mimicking their unnecessary role in LPCA-SRC on this database. The value $k=1$ was consistently selected by both $k$NN and $k$NN-Ext during cross-validation.

\textbf{Runtime results on Extended Yale B and ORL.} Tables \ref{Ext_Yale_B_time} and \ref{ORL_time} show the runtime and related results for the Extended Yale B and ORL experiments, respectively. LPCA-SRC had much longer runtimes than SRC on Extended Yale B, especially as $m_\mathrm{PCA}$ increased. This was due to a combination of large values for $d$ selected during cross-validation and the tangent vectors' decreasing efficacy at larger feature dimensions. However, the dictionary pruning procedure in LPCA-SRC actually eliminated a large number of training samples for all $m_\mathrm{PCA}$; once again, the computed tangent vectors contained more class-discriminating information than the eliminated nonlocal training samples, especially at lower feature dimensions for which details provided by these tangent vectors were especially needed. The linearity of the class manifolds of Extended Yale B, combined with this database's relatively dense sampling, lent itself well to the accurate computation of tangent vectors---part of the reason why LPCA-SRC used so many of them. Viewing these points as newly-generated and nearby training samples, LPCA-SRC's boost in accuracy over SRC can be viewed as an argument for locality in classification. We note that we might be able to decrease the value of $d$ in LPCA-SRC while still maintaining an advantage over SRC (see the discussion in Section \ref{sec:pars_dn}); our cross-validation procedure is designed to obtain the highest accuracy without regard to computational cost. 

On Extended Yale B, the TDC methods ran relatively more quickly (compared to the other algorithms) than on AR, presumably due to the much smaller number of classes on this database; both had runtimes typically between those of LPCA-SRC and SRC. Again, we see that LSDL-SRC had a relatively slow runtime for $m_\mathrm{PCA} = 30$ and became more competitive as $m_\mathrm{PCA}$ increased. Though both $k$NN and $k$NN-Ext were very fast, the large ``dictionary sizes'' in $k$NN-Ext made this algorithm clearly the slower of the two methods.

On ORL, LPCA-SRC and SRC had comparable runtimes, a result of rigorous dictionary pruning in LPCA-SRC. This algorithm and SRC$_\mathrm{pruned}$ retained roughly the same number of training samples in their respective dictionaries, and the latter was notably fast, running in about half the time as SRC. The remaining algorithms were even more efficient. TDC1 and TDC2 had comparable runtimes, both running faster than LSDL-SRC. As before, $k$NN and $k$NN-Ext had the fastest runtimes; the former was faster than the latter.

\subsubsection{Tangent vectors and PCA feature dimension} \label{sec:pca_tv}

In this section, we offer evidence to support our claim that the tangent vectors in LPCA-SRC can recover discriminative information lost during PCA transforms to low dimensions. Thus LPCA-SRC can offer a clear advantage over SRC in these cases, as we saw in experimental results on AR and Extended Yale B. 

\begin{figure}[H]
 \hspace*{\fill}
\centering
\begin{subfigure}[b]{0.16\textwidth}
\centering
	\includegraphics[width=\linewidth]{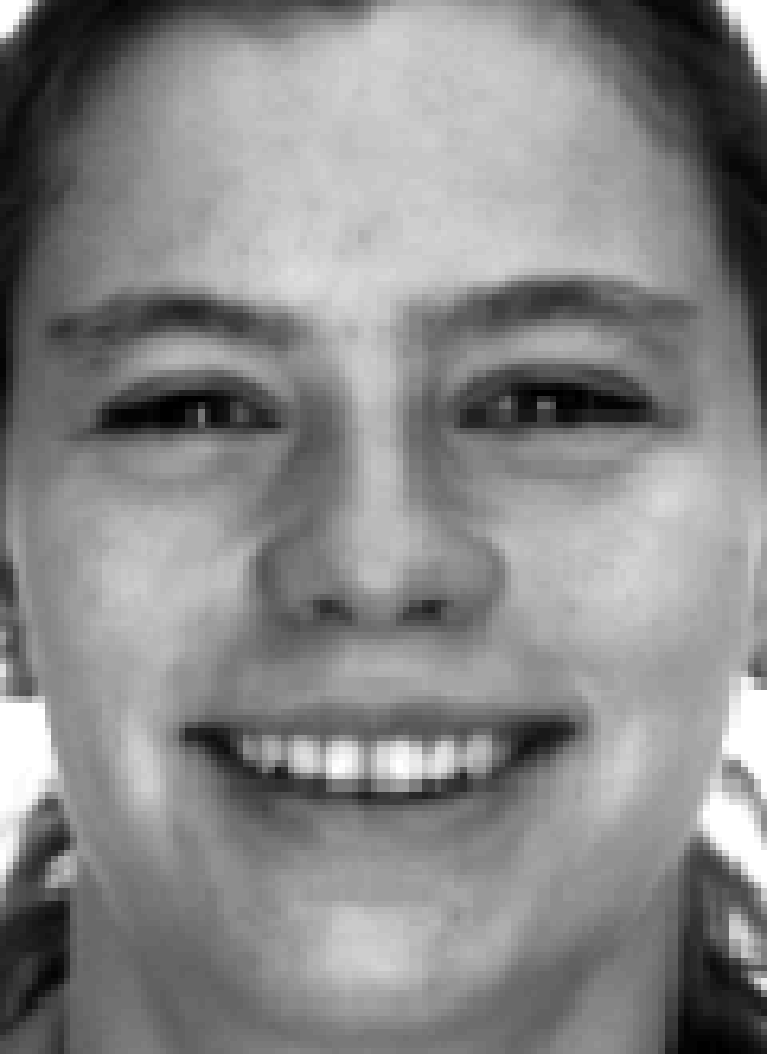}
		\caption{\scriptsize{Original Image}}
	\label{fig:orig_image_ex_2} \hfill
\end{subfigure}
\begin{subfigure}[b]{0.16\textwidth} 
\centering
	\includegraphics[width=\linewidth]{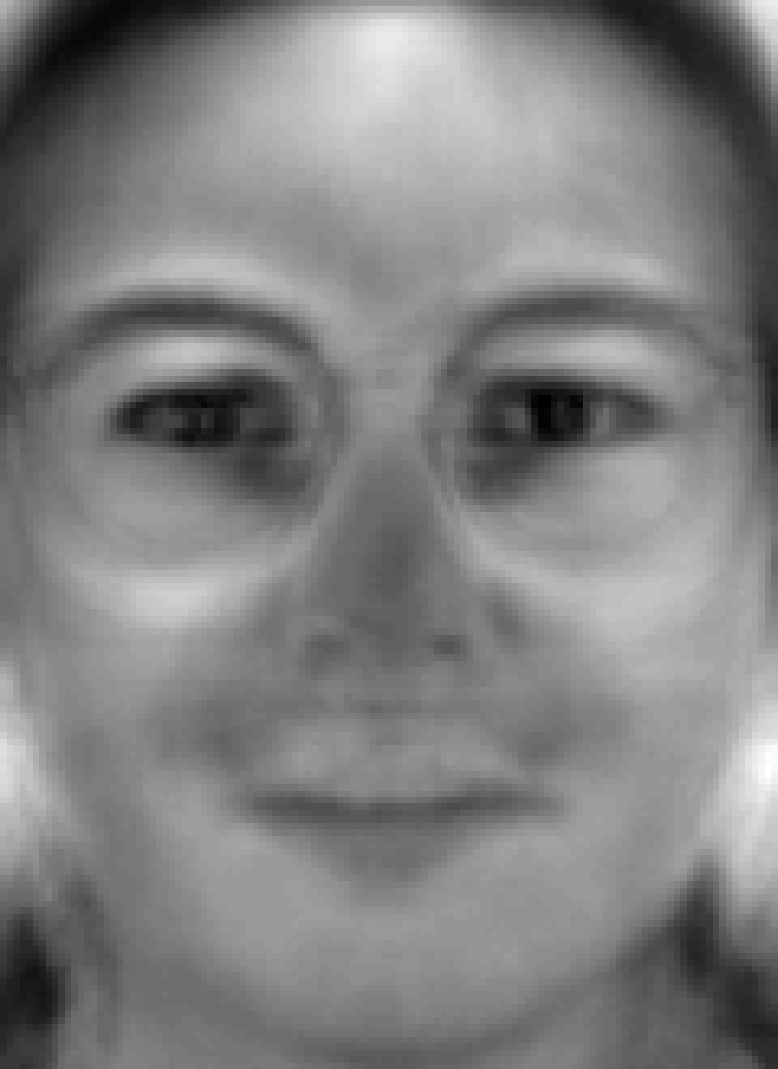}
		\caption{\scriptsize{Recovered Image}}
	\label{fig:recov_image_ex_2} \hfill
\end{subfigure}
\begin{subfigure}[b]{0.16\textwidth} 
\centering
	\includegraphics[width=\linewidth]{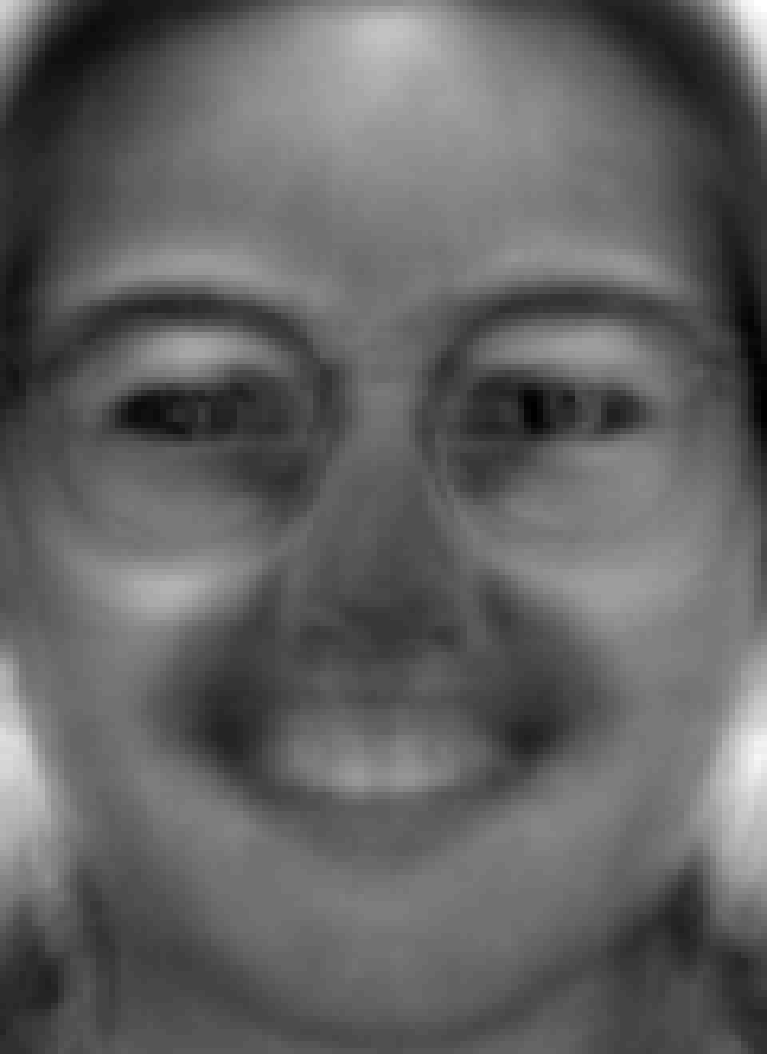}
		\caption{\scriptsize{Tangent Vector}}
	\label{fig:tv_ex_2}
	\hfill
\end{subfigure}
\hspace*{\fill}
\caption{The tangent vector does a much better job of displaying facial details conveying ``happiness'' than the recovered image. Images (b) and (c) were recovered from PCA dimension $m_\mathrm{PCA}=30$.}
\label{fig:AR_PCA_2}
\end{figure}

\begin{figure}[H]
 \hspace*{\fill}
\centering
\begin{subfigure}[b]{0.16\textwidth}
\centering
	\includegraphics[width=\linewidth]{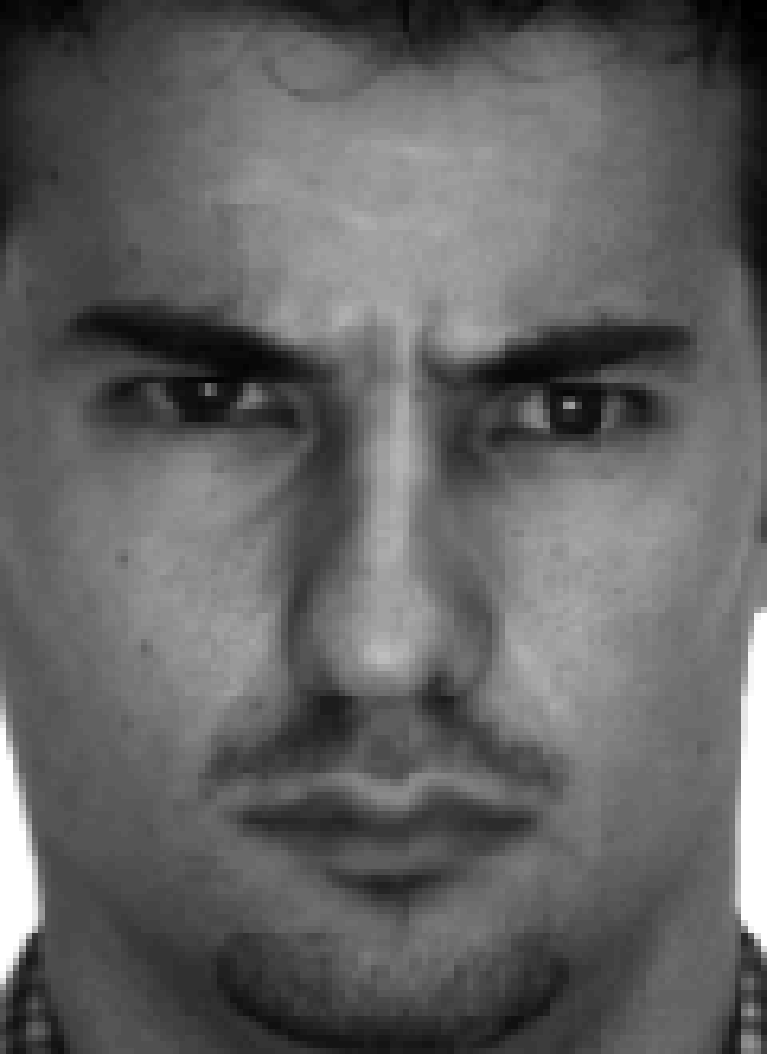}
		\caption{\scriptsize{Original Image}}
	\label{fig:orig_image_ex_3} \hfill
\end{subfigure}
\begin{subfigure}[b]{0.16\textwidth} 
\centering
	\includegraphics[width=\linewidth]{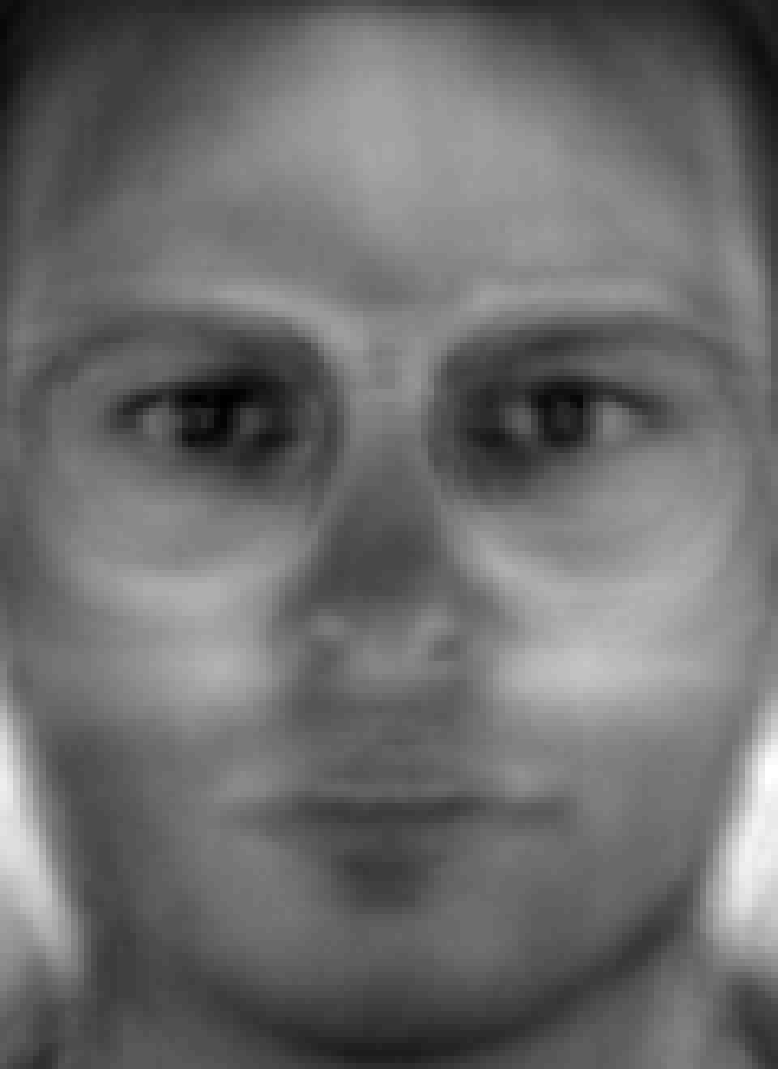}
		\caption{\scriptsize{Recovered Image}}
	\label{fig:recov_image_ex_3} \hfill
\end{subfigure}
\begin{subfigure}[b]{0.16\textwidth} 
\centering
	\includegraphics[width=\linewidth]{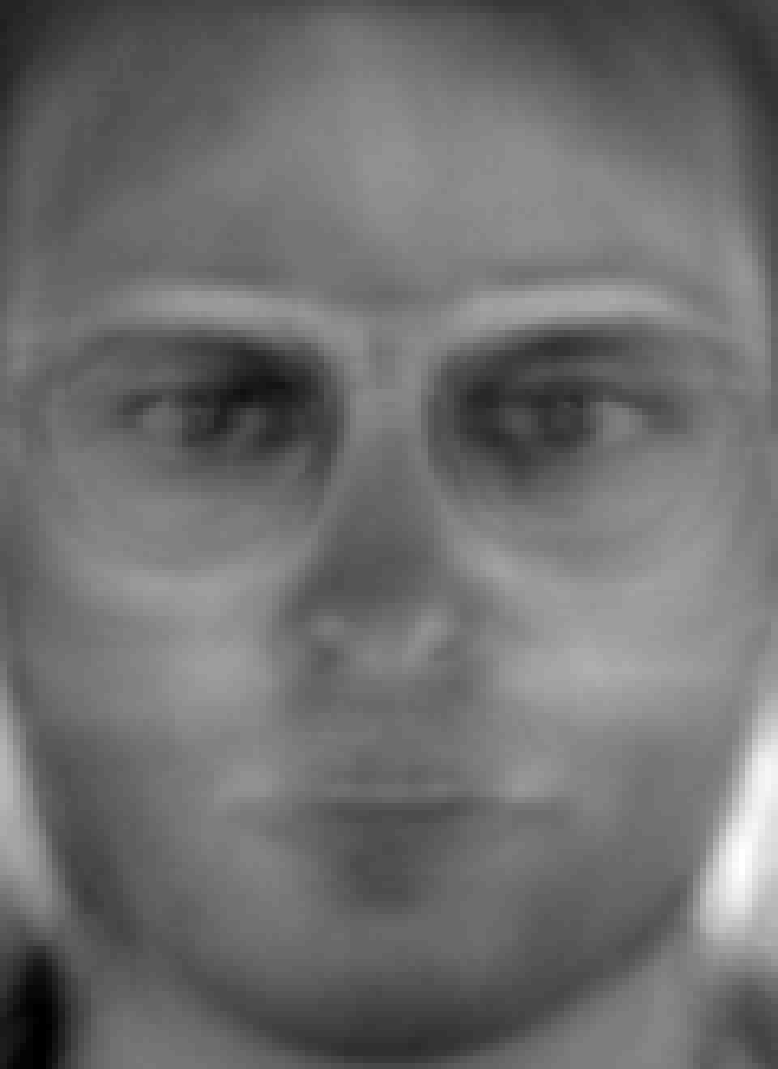}
		\caption{\scriptsize{Tangent Vector}}
	\label{fig:tv_ex_3}
	\hfill
\end{subfigure}
\hspace*{\fill}
\caption{The tangent vector does a better job of displaying ``anger'' than the recovered image, most notably in the subject's eyes and eyebrows. Images (b) and (c) were recovered from PCA dimension $m_\mathrm{PCA}=30$.}
\label{fig:AR_PCA_3}
\end{figure}

\begin{figure}[H]
 \hspace*{\fill}
\centering
\begin{subfigure}[b]{0.16\textwidth}
\centering
	\includegraphics[width=\linewidth]{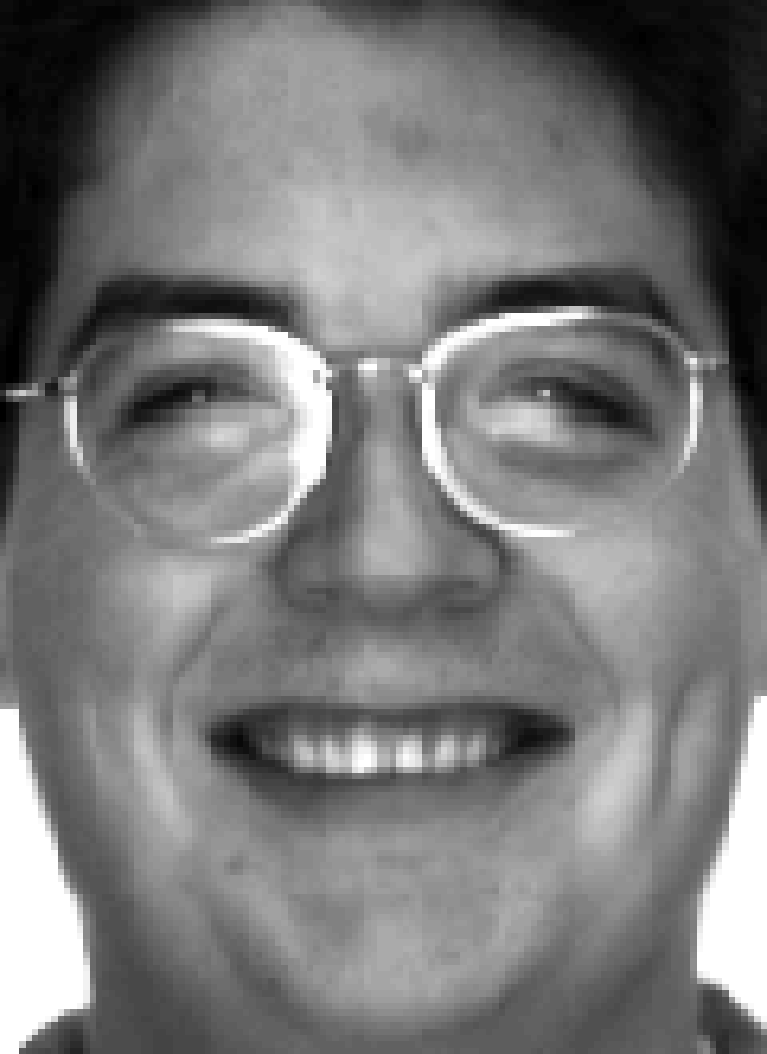}
		\caption{\scriptsize{Original Image}}
	\label{fig:orig_image_ex_4} \hfill
\end{subfigure}
\begin{subfigure}[b]{0.16\textwidth} 
\centering
	\includegraphics[width=\linewidth]{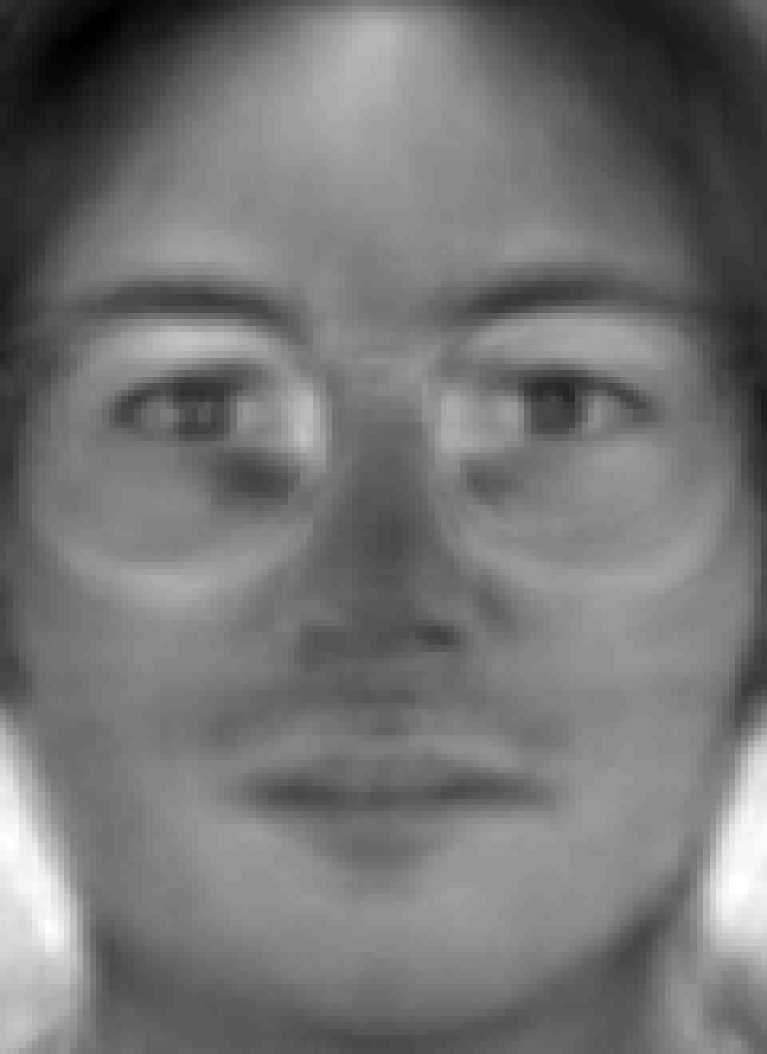}
		\caption{\scriptsize{Recovered Image}}
	\label{fig:recov_image_ex_4} \hfill
\end{subfigure}
\begin{subfigure}[b]{0.16\textwidth} 
\centering
	\includegraphics[width=\linewidth]{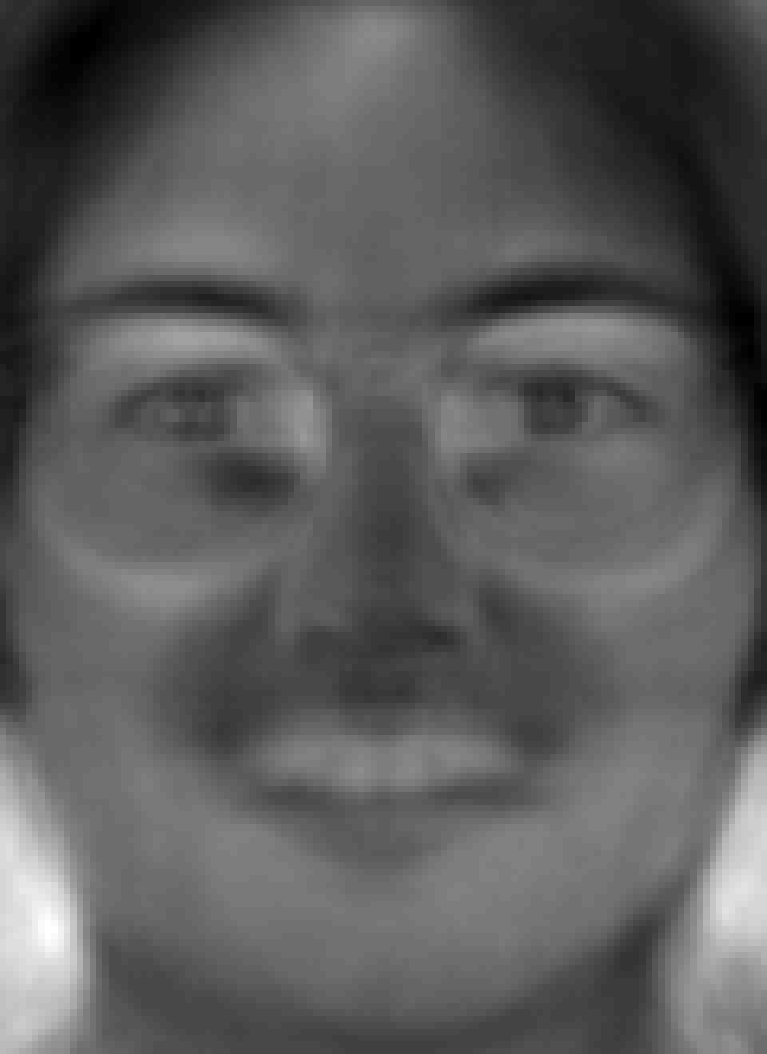}
		\caption{\scriptsize{Tangent Vector}}
	\label{fig:tv_ex_4}
	\hfill
\end{subfigure}
\hspace*{\fill}
\caption{The tangent vector shows the subject's smile better than the recovered image. Images (b) and (c) were recovered from PCA dimension $m_\mathrm{PCA}=30$. }
\label{fig:AR_PCA_4}
\end{figure}

In Figures \ref{fig:AR_PCA_2}-\ref{fig:AR_PCA_4}, we display three versions of three example images from AR-1. The first version is the original image (before PCA dimension reduction), the second version is the recovered image from PCA dimension reduction to dimension $m_\mathrm{PCA} = 30$, and the third version is the recovered corresponding tangent vector computed in LPCA-SRC. In each case, the tangent vector contains details of the original image not found in the recovered image, supporting our claim that the tangent vectors in LPCA-SRC can recover some (but not all) of the information lost in stringent PCA dimension reduction.

Towards quantifying what we mean by ``stringent,'' Table \ref{tab:energy} lists the average energy\footnote{By ``energy,'' we mean the ratio of the sum of squares of the first $m_\mathrm{PCA}$ singular values to the sum of squares of all singular values.} (over 10 trials) retained in the first $m_\mathrm{PCA}$ left-singular vectors of the face database training sets, along with the percent improvement in the accuracy of LPCA-SRC over that of SRC and SRC$_\mathrm{pruned}$. Given that the addition of tangent vectors did not increase classification accuracy on ORL, we see a correlation between the efficacy of tangent vectors in LPCA-SRC and the stringency of the PCA dimension reduction.

\begin{table*}[!htb]
\small{
\centering
\begin{tabular}{|c|c|c|c|c|c|c|}
\hline
& \multicolumn{2}{|c|}{$m_{\mathrm{PCA}} = 30$} & \multicolumn{2}{|c|}{$m_{\mathrm{PCA}} = 56$} & \multicolumn{2}{|c|}{$m_{\mathrm{PCA}} = 120$} \\
\hline
 & 	 & \% Increased Acc. & 	 & \% Increased Acc. & 	 & \% Increased Acc. \\
Database & Energy	 & SRC / SRC$_\mathrm{pruned}$ & Energy & SRC / SRC$_\mathrm{pruned}$ &	Energy & SRC / SRC$_\mathrm{pruned}$ \\
\hline
AR-1	&	0.4527	& 3.90/3.86 &	0.5322	& 1.87/1.91 &	0.6522 & 0.80/0.60	\\
AR-2 	&	0.4137	& 3.83/2.36 &	0.4884	& 1.31/0.63 &	0.5988 & 0.62/0.53	\\
Extended Yale B	&	0.3954	& 2.46/2.45 &	 0.4803 & 1.59/1.59 	&	0.6055 & 0.77/0.74	\\
ORL	&	0.5385	& 1.34/0.05 &	0.6581	& 1.26/-0.04 &	0.8487 & 1.73/0.03	\\
\hline
\end{tabular}
\caption{Average energy retained in PCA dimension reduction (over 10 trials) to various dimensions $m_\mathrm{PCA}$ on the face database training sets, as well as the average increase in classification accuracy of LPCA-SRC over SRC and SRC$_\mathrm{pruned}$. } 
\label{tab:energy}}
\end{table*}

\subsubsection{Summary} \label{sec:face_sum}

The experimental results on face databases show that LPCA-SRC can achieve higher accuracy than SRC in cases of low sampling and/or nonlinear class manifolds and small PCA feature dimension. We showed that LPCA-SRC had a significant advantage (in terms of mean accuracy) over SRC and the other algorithms for the small class sizes and nonlinear class manifolds of the AR database when the feature dimension was low. We also showed that LPCA-SRC could improve classification on Extended Yale B and ORL through its use of tangent vectors to provide a local approximation of the test sample and its discriminating pruning parameter, respectively. 

The runtime of LPCA-SRC was sometimes much longer than that of SRC, although this was less often seen for small feature dimensions, at which LPCA-SRC tended to excel. The size of the dictionary in LPCA-SRC was observed to be a good predictor of the relationship between the runtimes of LPCA-SRC and SRC, and this could easily be computed (given estimates of the parameters $n$ and $d$) before deciding between the two methods. 

To validate our claim that the tangent vectors in LPCA-SRC can contain information lost in stringent PCA dimension reduction, we provided examples from the AR database. We also compared the energy retained in PCA dimension reduction with the increase in accuracy in LPCA-SRC over SRC and saw that there was a correlation.

\section{Further Discussion and Future Work} \label{sec:fut_work}

This paper presented a modification of SRC called \emph{local principal component analysis SRC}, or ``LPCA-SRC.'' Through the use of tangent vectors, LPCA-SRC is designed to increase the sampling density of training sets and thus improve class discrimination on databases with sparsely sampled and/or nonlinear class manifolds. The LPCA-SRC algorithm computes basis vectors of approximate tangent hyperplanes at the training samples in each class and replaces the dictionary of training samples in SRC with a local dictionary (that is constructed based on each test sample) computed from shifted and scaled versions of these vectors and their corresponding training samples. Using a synthetic database and three face databases, we showed that LPCA-SRC can regularly achieve higher accuracy than SRC in cases of sparsely sampled and/or nonlinear class manifolds, low noise, and relatively small PCA feature dimension.

To address the issue of parameter setting, we recommended a consecutive parameter cross-validation procedure and gave detailed guidelines (including specific examples) for its use.
We also briefly discussed alternative methods for determining the class manifold dimension estimate $d$. It is important to note that in the case of small training sets, e.g., many face recognition problems, there are few options for the number-of-neighbors parameter $n$---and consequently for $d$ by Eq.~\eqref{eq:dn_range}---and so these values can easily be set using cross-validation, as in our experiments. When the training sets are very small (i.e., $N_l = 4$ or 5), one could simply set $n$ to its maximum value, i.e., $n=N_{l_\mathrm{min}}-2$, per Eq.~\eqref{eq:dn_range}. On the other hand, simply setting $d=1$ may suffice, especially when minimizing algorithm runtime and/or storage requirements is paramount.

One disadvantage of this method is its high computational cost and storage requirements. SRC is already expensive due to its $\ell^1$-minimization procedure; in LPCA-SRC, the computation of tangent vectors is added to the algorithm's workload. The size of the dictionary in LPCA-SRC may be larger or smaller than that of SRC, depending on the LPCA-SRC parameters $n$ and $d$ and the effect of the pruning parameter $r$. Thus LPCA-SRC can be slower or faster than SRC. Further, the storage required by LPCA-SRC is $(d+1)$ times that of SRC, which may be prohibitive when $d$ is large. As mentioned, simple computations based on the training set could render relative cost and storage estimates of using LPCA-SRC instead of SRC, and a smaller value of $d$ than that found using cross-validation (e.g., $d=1$) may be used successfully. These estimates can help the user decide between LPCA-SRC and SRC based on their desired balance between accuracy and computational efficiency. 

Additionally, as we saw on the synthetic database, the usefulness of the tangent vectors in LPCA-SRC decreases as the noise level in the training data increases. This problem could potentially be alleviated by using the method proposed by Kaslovsky and Meyer \cite{mey:tan} to estimate clean points on the manifolds from noisy samples and then computing the tangent vectors at these points. Note that the case of large training sample noise was the only case for which we saw LPCA-SRC not obtain higher accuracy than SRC. Thus LPCA-SRC should be preferred over SRC in low noise scenarios on either small-scale problems (e.g., the size of ORL) or when achieving a modest (e.g., $1\%-4\%$) boost in accuracy is worth potentially higher computational cost.

Open questions regarding LPCA-SRC include whether or not the aforementioned general trends hold for different methods of dimension reduction besides PCA. Additionally, one could compare the performance of the ``group'' or ``per-class'' methods of the above representation-based algorithms, in which test samples are approximated using class-specific dictionaries (similarly to as in TDC1). Lastly, one could gain insight into the role of $\ell^1$-minimization in SRC by comparing LPCA-SRC and SRC$_\mathrm{pruned}$ to versions of these algorithms that replace the $\ell^1$-norm with the $\ell^2$-norm, analogous to the work of Zhang et al.\ in their \emph{collaborative representation-based representation} model \cite{zha:crc2}. This is part of our ongoing work, which we hope to report at a later date.

\section*{Acknowledgments}

C. Weaver's research on this project was conducted with government support under contract FA9550-11-C-0028 and awarded by DoD, Air Force Office of Scientific Research, National Defense Science and Engineering Graduate (NDSEG) Fellowship, 32 CFR 168a. She was also supported by National Science Foundation VIGRE DMS-0636297 and NSF DMS-1418779. N. Saito was partially supported by ONR grants N00014-12-1-0177 and N00014-16-1-2255, as well as NSF DMS-1418779.


\bibliography{Bib_Master}

\appendix
\section{Tests of Statistical Significance}\label{sec:appendix}

This appendix contains the detailed results for the tests of statistical significance between the most competitive classification algorithms on the experiments presented in Sections \ref{sec:syn_exps}, \ref{sec:AR}, and \ref{sec:yale_orl}. 

\subsection{Tests of Statistical Significance for Experiments on the Synthetic Database}

Recall that LPCA-SRC, SRC, and SRC$_\mathrm{pruned}$ were the most competitive algorithms on the synthetic database experiments presented in Section \ref{sec:syn_exps}. As evidence that LPCA-SRC outperformed SRC in a statistically-significant manner, we performed a Repeated Measures ANOVA test on all three methods as well as a t-test between the results for LPCA-SRC and SRC. The corresponding $p$-values and confidence intervals are contained in Tables \ref{table:anova_syn_N0} and \ref{table:anova_syn_eta}. The columns of these tables are as follows: The value of $N_0$ (in the case of varying class size) or $\eta$ (in the case of varying noise level) in the experiment, the $p$-value for Univariate Type III Repeated-Measures ANOVA Assuming Sphericity, the $p$-value for Mauchly Tests for Sphericity, the $p$-values for Greenhouse-Geisser and Huynh-Feldt Corrections for Departure from Sphericity, and the 5\% confidence interval for a one-sided t-test of the improvement of LPCA-SRC over SRC. These tests were performed in \texttt{R} with the functions \texttt{Anova} (from the car package) and \texttt{t.test}. 

For all but the Mauchly test, a small $p$-value indicates that we should reject the null hypothesis, which states that the algorithms have the same average accuracy. For the Mauchly test, a large $p$-value indicates that the data obeys the sphericity assumption; otherwise, the Greenhouse-Geisser or Huynh-Feldt corrections should be used. The confidence intervals can be interpreted as follows: Were we to repeat this experiment, we would expect LPCA-SRC to outperform SRC (with the exception of $N_0=15$ and $\eta\geq 0.01$) with the difference in mean accuracies falling within this confidence interval 95 times out of 100.

\begin{table*}[!htb] 
\small{
\centering
\begin{tabular}{|c|c|c|c|c|c|}
\hline
$N_0$ & rANOVA & Mauchly & Greenhouse-Geisser & Huynh-Feldt & 5\% Confidence (LPCA-SRC $>$ SRC)  \\
\hline
5 & $4.1 \times 10^{-5}$  & $0.3598$ & $4.7 \times 10^{-5}$ & $4.1 \times 10^{-5} $& $[0.0266, 0.0804]$\\
10 & $2.0 \times 10^{-9}$  & $3.9 \times 10^{-9}$ & $1.2 \times 10^{-7}$ & $1.0 \times 10^{-7} $& $[0.0276, 0.0604]$\\
15 & $0.1488$  & $2.3 \times 10^{-10}$ & $0.1610$ & $0.1606 $& $[-0.0025, 0.0265]$\\
20 & $5.8 \times 10^{-10}$  & $1.1 \times 10^{-8}$ & $3.9 \times 10^{-8}$ & $3.3 \times 10^{-8} $& $[0.0244, 0.0526]$\\
25 & $1.7 \times 10^{-8}$  & $1.1 \times 10^{-18}$ & $3.3 \times 10^{-6}$ & $3.1 \times 10^{-6} $& $[0.0166, 0.0448]$\\
30 & $2.2 \times 10^{-16}$  & $1.1 \times 10^{-13}$ & $9.6 \times 10^{-14}$ & $7.5 \times 10^{-14} $& $[0.0349, 0.0611]$\\
35 & $2.2 \times 10^{-16}$  & $7.0 \times 10^{-10}$ & $4.8 \times 10^{-13}$ & $3.6 \times 10^{-13} $& $[0.0309, 0.0540]$\\
40 & $1.1 \times 10^{-6}$  & $5.2 \times 10^{-23}$ & $7.1 \times 10^{-5}$ & $6.8 \times 10^{-5} $& $[0.0113, 0.0361]$\\
45 & $2.2 \times 10^{-16}$  & $8.6 \times 10^{-8}$ & $2.2 \times 10^{-16}$ & $6.0 \times 10^{-18} $& $[0.0248, 0.0395]$\\
50 & $2.2 \times 10^{-16}$  & $1.9 \times 10^{-17}$ & $2.2 \times 10^{-16}$ & $2.0 \times 10^{-31} $& $[0.0331, 0.0439]$\\
55 & $2.2 \times 10^{-16}$  & $1.9 \times 10^{-13}$ & $2.2 \times 10^{-16}$ & $5.0 \times 10^{-37} $& $[0.0330, 0.0423]$\\
60 & $2.2 \times 10^{-16}$  & $8.4 \times 10^{-5}$ & $2.2 \times 10^{-16}$ & $3.4 \times 10^{-33} $& $[0.0258, 0.0356]$\\
65 & $2.2 \times 10^{-16}$  & $3.2 \times 10^{-10}$ & $2.2 \times 10^{-16}$ & $1.0 \times 10^{-41} $& $[0.0285, 0.0356]$\\
70 & $2.2 \times 10^{-16}$  & $3.6 \times 10^{-20}$ & $2.2 \times 10^{-16}$ & $6.1 \times 10^{-38} $& $[0.0291, 0.0366]$\\
75 & $2.2 \times 10^{-16}$  & $7.8 \times 10^{-23}$ & $2.2 \times 10^{-16}$ & $1.7 \times 10^{-31} $& $[0.0264, 0.0343]$\\
\hline
\end{tabular}
\caption{Tests for statistical significance on the synthetic database for varying $N_0$: $p$-values in rAnova, Mauchly, Greenhouse-Geisser, and Huynh-Feldt tests (LPCA-SRC, SRC, and SRC$_\mathrm{pruned}$), and 5\% confidence interval of the improvement of LPCA-SRC over SRC.}
\label{table:anova_syn_N0}}
\end{table*}

\begin{table*}[!htb] 
\small{
\centering
\begin{tabular}{|c|c|c|c|c|c|}
\hline
$\eta$ & rANOVA & Mauchly & Greenhouse-Geisser & Huynh-Feldt & Confidence (LPCA-SRC $>$ SRC)  \\
\hline
0.0001 & $2.2 \times 10^{-16}$  & $8.1 \times 10^{-21}$ & $6.4 \times 10^{-14}$ & $5.4 \times 10^{-14} $& $[0.0356, 0.0596]$\\
0.001 & $2.3 \times 10^{-6}$  & $9.5 \times 10^{-10}$ & $2.9 \times 10^{-5}$ & $2.7 \times 10^{-5} $& $[0.0135, 0.0428]$\\
0.005 & $1.4 \times 10^{-9}$  & $5.1 \times 10^{-7}$ & $4.6 \times 10^{-8}$ & $3.8 \times 10^{-8} $& $[0.0232, 0.0529]$\\
0.01 & $0.0938$  & $1.3 \times 10^{-21}$ & $0.1180$ & $0.1177 $& $[-0.0073, 0.0271]$\\
0.015 & $0.9044$  & $9.6 \times 10^{-13}$ & $0.8325$ & $0.8345 $& $[-0.0147, 0.0110]$\\
0.02 & $0.0027$  & $2.8 \times 10^{-19}$ & $0.0098$ & $0.0096 $& $[-0.0334,-0.0064]$\\
0.03 & $5.3 \times 10^{-5}$  & $9.0 \times 10^{-17}$ & $0.0006$ & $0.0006 $& $[-0.0388,-0.0109]$\\
0.05 & $0.0004$  & $9.0 \times 10^{-9}$ & $0.0014$ & $0.0013 $& $[-0.0254,-0.0051]$\\
\hline
\end{tabular}
\caption{Tests for statistical significance on the synthetic database for varying $\eta$: $p$-values in rAnova, Mauchly, Greenhouse-Geisser, and Huynh-Feldt tests (LPCA-SRC, SRC, and SRC$_\mathrm{pruned}$), and 5\% confidence interval of the improvement of LPCA-SRC over SRC.}
\label{table:anova_syn_eta}}
\end{table*}

\subsection{Tests of Statistical Significance for Experiments on the Face Databases}

To test for statistical significance in the differences between algorithm accuracy on the AR, Extended Yale B, and ORL face databases, we performed Repeated Measures ANOVA tests on LPCA-SRC, SRC, SRC$_\mathrm{pruned}$, and LSDL-SRC as well as two t-tests on each database, one between LPCA-SRC and SRC and the other between LPCA-SRC and LSDL-SRC. The related $p$-values and confidence intervals are contained in Tables \ref{table:anova_face_AR} and \ref{table:anova_face_yale_orl}. The columns in these tables are as follows: The name of the database, the PCA dimension $m_\mathrm{PCA}$, the $p$-value for Univariate Type III Repeated-Measures ANOVA Assuming Sphericity, the $p$-value for Mauchly Tests for Sphericity, the $p$-values for Greenhouse-Geisser and Huynh-Feldt Corrections for Departure from Sphericity, the 5\% confidence interval for a one-sided t-test between LPCA-SRC and SRC (LPCA-SRC $>$ SRC), and the 5\% confidence interval for a one-sided t-test between LPCA-SRC and LSDL-SRC (LPCA-SRC $>$ LSDL-SRC). For all but the Mauchly test, a small $p$-value indicates that we should reject the null hypothesis, which states that the algorithms have the same average accuracy. For the Mauchly test, a large $p$-value indicates that the data obeys the sphericity assumption; otherwise, the Greenhouse-Geisser or Huynh-Feldt corrections should be used. These tests were performed in \texttt{R} with the functions \texttt{Anova} (from the car package) and \texttt{t.test}. 

\begin{table*}[!htb] 
\small{
\centering
\begin{tabular}{|c|c|c|c|c|c|c|c|}
\hline
Database & $m_\mathrm{PCA}$ & rANOVA & Mauchly & G-G & H-F & t-test$_1$ & t-test$_2$ \\
\hline
AR-1 & 30 & $8.5 \times 10^{-12}$  & $0.9182$ & $9.1 \times 10^{-11}$ & $8.5 \times 10^{-12} $& $[0.0310, 0.0499]$ & $[0.0389, 0.0567]$\\
AR-1 & 56 & $8.0 \times 10^{-7}$  & $0.2124$ & $0.0001$ & $1.8 \times 10^{-5} $& $[0.0145,0.0226]$ & $[0.0076,0.0165]$\\
AR-1 & 120 & $3.7 \times 10^{-7}$  & $0.0470$ & $7.3 \times 10^{-5}$ & $1.3 \times 10^{-5} $& $[0.0039, 0.0117]$ & $[-0.0087,-0.0018]$\\
AR-2 & 30 & $1.3 \times 10^{-10}$  & $0.0083$ & $5.0 \times 10^{-7}$ & $2.2 \times 10^{-8} $& $[0.0243,0.0497]$ & $[0.0544, 0.0952]$\\
AR-2 & 56 & $1.8 \times 10^{-5}$  & $0.0651$ & $0.0009$ & $0.0003 $& $[0.0047,0.0222]$ & $[0.0139,0.0286]$\\
AR-2 & 120 & $0.0115$  & $0.4566$ & $0.0225$ & $0.0115 $& $[0.0009,0.0095]$ & $[-0.0046,0.0055]$\\
\hline
\end{tabular}
\caption{Tests for statistical significance on the AR Face Database: $p$-values in rAnova, Mauchly, Greenhouse-Geisser, and Huynh-Feldt tests (LPCA-SRC, SRC, SRC$_\mathrm{pruned}$, and LSDL-SRC), and 5\% confidence intervals of the improvement of LPCA-SRC over SRC (t-test$_1$) and over LSDL-SRC (t-test$_2$).}
\label{table:anova_face_AR}}
\end{table*}

\begin{table*}[!htb] 
\small{
\centering
\begin{tabular}{|c|c|c|c|c|c|c|c|}
\hline
Database & $m_\mathrm{PCA}$ & rANOVA & Mauchly & G-G & H-F & t-test$_1$ & t-test$_2$ \\
\hline
Yale B & 30 & $2.2 \times 10^{-16}$  & $0.0014$ & $1.0 \times 10^{-11}$ & $4.8 \times 10^{-13} $& [0.0182,0.0300] & $[0.1422,0.1687]$\\
Yale B & 56 & $2.2 \times 10^{-16}$  & $4.2 \times 10^{-5}$ & $4.6 \times 10^{-5}$ & $8.4 \times 10^{-12} $& [0.0088,0.0233] & $[0.0671,0.0822]$\\
Yale B & 120 & $9.1 \times 10^{-14}$  & $0.1026$ & $1.5 \times 10^{-9}$ & $6.3 \times 10^{-12} $& [0.0052,0.0102] & $[0.0190,0.0192]$\\
ORL & 30 & $2.2 \times 10^{-16}$  & $0.0033$ & $3.9 \times 10^{-12}$ & $1.1 \times 10^{-12} $& [0.0085,0.0168] & $[0.0092,0.0192]$\\
ORL & 56 & $4.3 \times 10^{-14}$  & $0.0012$ & $5.7 \times 10^{-12}$ & $1.6 \times 10^{-12} $& $[0.0121,0.0201]$ & $[0.0045,0.0120]$\\
ORL & 120 & $2.2 \times 10^{-16}$  & $3.2 \times 10^{-5}$ & $2.2 \times 10^{-16}$ & $7.8 \times 10^{-35} $ & $[0.0142,0.0213]$ & $[0.0308,0.0400]$\\
\hline
\end{tabular}
\caption{Tests for statistical significance on the Extended Yale B and ORL Face Database: $p$-values in rAnova, Mauchly, Greenhouse-Geisser, and Huynh-Feldt tests (LPCA-SRC, SRC, SRC$_\mathrm{pruned}$, and LSDL-SRC), and 5\% confidence intervals of the improvement of LPCA-SRC over SRC (t-test$_1$) and over LSDL-SRC (t-test$_2$).}
\label{table:anova_face_yale_orl}}
\end{table*}

\end{document}